\documentclass{bmvc2k}

\title{Learning Neural Transmittance for Efficient Rendering of Reflectance Fields}

\addauthor{Mohammad Shafiei}{moshafie@ucsd.edu}{1}
\addauthor{Sai Bi}{bisai@cs.ucsd.edu}{{1,2}}
\addauthor{Zhengqin Li}{zhl378@ucsd.edu}{1}
\addauthor{Aidas Liaudanskas}{aliaudanskas@fyusion.com}{3}
\addauthor{Rodrigo Ortiz-Cayon}{rcayon@fyusion.com}{3}
\addauthor{Ravi Ramamoorthi}{ravir@ucsd.edu}{1}

\addinstitution{
 University of California San Diego
}
\addinstitution{
 Adobe Research
}
\addinstitution{
 Fyusion Inc.
}

\runninghead{M. Shafiei, S. Bi, et al.}{Neural Transmittance for Efficient Rendering}

\usepackage{caption}[skip=0pt]
\usepackage{multirow}
\usepackage{ctable} 
\usepackage{bm}
\usepackage{siunitx}
\usepackage{tabularx}
\usepackage{float}
\usepackage{dcolumn}

\newcolumntype{Y}{D..{6.4}}

\newcommand{\rr}[1]{{\color{red}{Ravi:#1}}}
\newcommand{\ms}[1]{{\color[rgb]{0,0.8,0}{Mohammad:#1}}}
\newcommand{\sai}[1]{{\color{purple}{[Sai:#1]}}}
\newcommand{\zl}[1]{{\color{red}{Zhengqin:#1}}}

\newcommand{\rebuttal}[1]{#1}

\newcommand{\boldstart}[1]{\noindent\textbf{#1}}
\newcommand{\boldstartspace}[1]{\vspace{0.1in}\noindent\textbf{#1}}
\newcommand{\cmnt}[1]{}

\newcommand{\speedupPointlight}{2}

\newcommand{\speedupFiveHunred}{92}
\newcommand{\speedupTransmittance}{380}

\usepackage[ruled,vlined]{algorithm2e}

\begin{document}

\maketitle

\begin{abstract}
   Recently neural volumetric representations such as neural reflectance fields 
   have been widely applied to faithfully reproduce the appearance of real-world objects and scenes under novel viewpoints and lighting conditions. 
   However, it remains challenging and time-consuming to render 
   such representations under complex lighting such as environment maps, 
   which requires individual ray marching towards each single light to
   calculate the transmittance at every sampled point. In this paper, we propose 
   a novel method based on precomputed Neural Transmittance Functions 
   to accelerate the rendering of neural reflectance fields. 
   Our neural transmittance functions enable us to efficiently query the transmittance 
   at an arbitrary point in space along an arbitrary ray  without tedious ray 
   marching, which effectively reduces the time-complexity of the rendering. 
   We propose a novel formulation for the neural transmittance function, and train it jointly with the neural reflectance fields on images captured under collocated camera and light, 
   while enforcing monotonicity.
   Results on real and synthetic scenes demonstrate almost
   two order of magnitude speedup for renderings under environment maps with minimal accuracy loss. 
 
\end{abstract}

\section{Introduction}


Reproducing the appearance of real-world objects and scenes is challenging. 
Traditional methods~\cite{nam2018practical,bi2020deep,zhou2016sparse,Newcombe2011kinectFusion} 
recover the reflectance of the scenes and geometry, represented with triangle meshes, 
to synthesize images under novel conditions. These methods often
fail to handle challenging scenes such as those with thin structures and  
severe occlusions.
Neural rendering methods such as neural 
radiance fields (NeRF)~\cite{mildenhall2020nerf} exploit volumetric scene 
representations and differentiable ray marching to
significantly improve the image quality. The original NeRF is limited
to pure view synthesis. Thus, recent works~\cite{bi2020neural,nerv2020}
extend the volumetric rendering framework 
and recover neural reflectance fields (NRF) to support rendering under both novel
viewpoints and lighting conditions. They consider light transport and recover the
spatial reflectance properties and volume density information. 

NRF faithfully reproduces the appearance of objects,
but it is time-consuming to render them under complex lighting conditions e.g. 
environment lighting. Specifically, assume a single-scattering model. 
For each sampled point on the camera ray, we need to evaluate the transmittance 
for each light in the scene, which involves another ray marching towards it. Moreover, 
each sample on the light ray requires an additional network inference.  
\cmnt{Previous methods such as Bi et al.~\cite{bi2020neural} accelerate the rendering by precomputing 
a transmittance volume over the light frustum.  During testing, they directly sample the 
precomputed transmittance volume to get the light transmittance of the desired point. Such 
a volume has a limited resolution and it is time-consuming and memory-intensive to 
build and store, especially when there are many lights in the scene.}


In this paper, we significantly reduce the time-complexity of the rendering by jointly learning
a {\em Neural Transmittance} function, along with other NRF parameters while training. 
Given such a function, we can directly query the transmittance of a point along an arbitrary 
direction without ray marching. 
The Neural transmittance function is modeled by a Multilayer Perceptron (MLP).
In its simplest form, the network could take the position of the point and the desired direction  
as input, and output the transmittance. However, such a simple formulation does not
account for physical priors on the transmittance function such as its monotonicity
along rays. Therefore, instead of predicting the transmittance for each sampled point
independently, we directly predict the transmittance function for the desired ray.

Our method is based on NRF~\cite{bi2020neural}, which 
applies an MLP to predict the volume density, reflectance, and normal of a point.
Similarly, we train our model on multi-view images captured with collocated camera and light. We 
jointly train the networks for NRF and our neural transmittance 
function, where we use the transmittance calculated with the former model to supervise the 
training of the latter model. Since the camera/light rays corresponding to pixels on training 
images only cover a small portion of the 
rays that pass the objects, simply training the neural transmittance networks on 
the rays corresponding to training pixels results in poor generalization to unseen rays.
Hence, we introduce an effective data augmentation by randomly 
sampling rays in space and calculating their transmittance to supervise the training of our 
neural transmittance networks. In Section~\ref{sec:results}, we show that our data augmentation 
effectively improves the quality of the results.



In summary, our main contributions are:
\begin{itemize}
    \setlength\itemsep{0.1pt}
\item 
 We propose to jointly train a novel neural transmittance network that enables us 
to effectively query the transmittance of a point along an arbitrary direction. 

\item 
    We apply a novel generalized logistic function to model the 
    monotonic transmittance along 
    the ray and use the network to predict the parameters for the logistic function 
    with a two-sphere parameterization.

\item We introduce a novel data augmentation method to provide additional supervisions for 
    the training of our 
    neural transmittance networks, which allows our network to better generalize 
    to unseen rays and achieve renderings of higher quality.


\item{We demonstrate that our method achieves almost two orders of magnitude speedup 
for rendering under complex lighting conditions such as environment lighting.}


\end{itemize}



\section{Related Work}

\boldstart{Geometry and reflectance acquistion.}
To reproduce the appearance of real-world objects, 
traditional methods~\cite{schoenberger2016sfm,nam2018practical,zhou2016sparse,bi2020deep} apply 
multi-view stereo to find correspondence across input images to reconstruct the geometry 
of the objects, usually represented with a triangle mesh. They estimate the material
properties of the object, commonly represented by SVBRDFs,  
by optimization to match the appearance of rendered images to the 
captured images. 
\cmnt{
While triangle meshes with per-vertex BRDFs can be efficiently rendered with modern 
rendering engines,  the problem is that it is typically challenging 
to reconstruct highly accurate triangle meshes for objects with complex geometry
such as thin structures and heavy occlusions, which greatly limits the quality 
of the reconstructions.} In contrast, we learn a volumetric
representation for the objects and propose an efficient method based on 
learnt precomputed transmittance for efficient rendering.

\boldstartspace{Neural representations.} Recent works exploit neural representations
to reproduce the appearance of scenes, which involves using neural networks to learn 
scene geometry and reflectance. Some methods directly predict the explicit 
geometry with representations such as 
point clouds~\cite{aliev2019neural}, occupancy grids~\cite{kar2017learning,sitzmann2019deepvoxels} 
and signed distance fields~\cite{liao2018deep,saito2019pifu}.
Volumetric representations~\cite{sitzmann2019scene,lombardi2019neural,bi2020DRV} are 
also applied to acquire the appearance of objects and scenes, where ray marching is performed to 
render the desired images. \cmnt{Compared to the other representations, volumetric representations
can better make use of the 3D space and generate renderings of higher quality. However, 
the explicit 3D volume requires a discretization of the space and its resolution is 
typically limited due to the constraint of GPU memory, hence limiting the quality 
of the rendered images. To solve this problem, }Mildenhall et al.~\cite{mildenhall2020nerf} 
propose to use an implicit representation for the volume where an MLP is used 
to predict the volume density and radiance of an arbitrary point by taking its 3D location  
and view direction as input. Bi et al.~\cite{bi2020neural} further proposes to jointly 
learn volume density and reflectance from flash images captured by mobile phones, which supports 
joint view synthesis and relighting. 
\cmnt{
While implicit representation effectively increases the resolution of the volume, it 
also significantly increases the computational cost. For each sampled point, we need to 
evaluate the network to get its properties. Such a problem is more intractable when it comes 
to relighting as done in~\cite{bi2020neural}, where the light transmittance for each 
sampled point needs to be calculated with another ray marching towards 
the light and integrating over all samples on the light ray. 
We propose to jointly train a neural transmittance network to directly predict 
the transmittance along the ray for efficient rendering. 
}
  
\boldstart{Precomputation techniques.}
Precomputing scene components such as radiance transfer~\cite{sloan2002precomputed,ng2003all} 
global illumination~\cite{ritschel2009micro,dong2007interactive} and 
visibility~\cite{lokovic2000deep,kallweit2017deep,nerv2020, bi2020neural} is used
to accelerate the rendering. Lokovic and Veach~\cite{lokovic2000deep} precompute a deep shadow 
map for efficient visibility lookup and high-quality shadows. Kallweit et al.~\cite{kallweit2017deep} 
learn a neural network to fit the in-scattered radiance for an arbitrary point and view direction.
Bi et al.~\cite{bi2020neural} precompute a transmittance volume at the light by calculating 
the transmittance of each sampled point on the rays corresponding to the 
pixels on the virtual image plane placed at the light. 

The work concurrent to ours by Srinivasan et al.~\cite{nerv2020}
predicts the transmittance with a network by taking the position of the 3D point and 
the lighting direction, which fails to conform to physical monotonicity of 
the transmittance. 
\cmnt{
Instead, we propose to fit the transmittance along the light ray with a  
generalized logistic function whose parameters are predicted with our networks.
    Such a design enables us to efficiently precompute a transmittance map 
    at the light by predicting the transmittance function of each ray 
    for the pixels on the virtual image plane at the light. Afterwards, the 
    transmittance of each point can be simply calculated using  
    the transmittance function of its corresponding ray.} 
    Compared to Bi et al.~\cite{bi2020neural} and Srinivasan et al.~\cite{nerv2020} 
    that predict the transmittance for each sample on the ray,   
    our method achieves significant speedups 
    while maintaining high-quality renderings.

\cmnt{
In contrast, Bi et al.  \cite{bi2020neural} require computation of a
discretized transmittance volume per light sample at test
time \cite{lokovic2000deep}\rr{Provide citations for transmittance volumes}. Neural
transmittance requires more than 2 gigabytes less memory compared to an explicit
opacity or transmittance volume \rr{Provide numbers}, which allows faster integration
of shadows directly in the rendering pass on the GPU \ms{on a similar device we have similar
time-complexity. Without downloading overhear we are faster.} \rr{on the GPU?}. 
In contrast, precomputed discretized transmittance
volumes as in~\cite{bi2020neural} require downloading the radiance and
transmittance on all samples of the volume to the CPU before
integrating them along the view direction. This would make the
relighting with environment map particularly time consuming as it
requires downloading the scene information per light sample. Our
method additionally enables continuous and independent sampling of
transmittance per sample point which allows Multiple Importance
Sampling (MIS).
}

\section{Background}
\label{sec:background}

Our method builds on the framework by Bi et al.~\cite{bi2020neural} that estimates 
NRF~\cite{bi2020neural} for joint view synthesis and relighting. 
They use MLPs to predict the reflectance, normal and volume density of points in the scene.
They apply differentiable volume rendering via ray marching to 
render the image. At each sample point on the ray, they determine its
contribution with a differentiable reflectance model, which integrates
to give the color of that pixel:
\begin{align}
\bm{x} &= \bm{r}(t)    = \bm{o} + \bm{\omega}_o t \\
\bm{L}(\bm r)       &= \int_0^\infty \tau(\bm{o}, \bm{x}, \bm{\omega}_o)\sigma(\bm{x})\bm{h}(\bm{x},\bm{\omega}_o)dt \label{eq:nerf} \\
\bm{h}(\bm{x}, \bm{\omega}_o)                         &= \int_\Omega \tau(\bm{l},\bm{x},\bm{\omega}_i)
\bm{\rho}(\bm{R}(\bm{x}),\bm{\omega}_o,\bm{\omega}_i,\bm{n}(\bm{x})) d\bm{\omega}_i \label{eq:nrf} \\
\tau(\bm{l}, \bm{x}, \bm{\omega})         &= \exp(-\int_0^{||\bm{l-x}||}\sigma(\bm{l} + u \bm{\omega}) du) 
\label{eq:transmittance}
\end{align}
where $\bm{x}$ is a 3D point that lies on the ray $\bm{r}$ at a
distance $t$ from camera origin $\bm{o}$ on ray direction $\bm{\omega}_o$.
Incident radiance $\bm{L}$ of that ray on the image plane is computed by an integral on the ray.
The integrand is a product of transmittance $\tau$ along view direction, volume
density $\sigma$ and outgoing radiance $\bm{h}$. Outgoing radiance is an integral
over the product of transmittance along light direction $\bm{\omega}_i$, a
differentiable reflectance model $\bm{\rho}$ that is a function of reflectance parameters $\bm{R}$,
incoming and outgoing directions and normal vector $\bm{n}$ on the point $\bm{x}$.
This integral is over the domain of upper hemisphere $\Omega$ to the normal.
$\tau$ is a function of light source location $\bm{l}$, $\bm{\omega}_i$
and $\bm{x}$. Boldface notations in this paper represent vectors.

Note that in this case evaluating the integral in Equation~\ref{eq:nerf} involves 
a double integral, where we integrate over multiple samples along the camera ray 
and the outgoing radiance of each sample is determined by an integral over the upper
hemisphere at its local surface. 
During training,
with the assumption of a single point light collocated with the camera
and a single scattering model,
the equation above can be simplified significantly, i.e., 
the integral in equation~\ref{eq:nrf} is removed,
and the transmittance for the light ray and camera ray would be identical, 
so only a single evaluation is needed.  
While this assumption alleviates the complicated integral at training
time, relighting the scene at test time is still computationally
intensive.
Specifically, to render the scene under uncollocated camera and light,
a na\"ive rendering process would still need to evaluate a double integral: 
for each sample on the camera ray, we need to perform another raymarching 
towards the light to evaluate the light transmittance, where the time-complexity
is quadratic in the number of samples.

Bi et al. \cite{bi2020neural} address this problem by precomputing a
transmittance volume inspired by Lokovic and Veach
\cite{lokovic2000deep}.  They place a virtual image plane at the light
and march a ray through each pixel and calculate the transmittance of 
each sample point on the ray, which effectively forms a 3D transmittance
volume. During testing, they directly query the transmittance of 
the desired point by interpolating the precomputed transmittance volume.  
This strategy reduces the number of network inferences for light transmittance 
to be linear in the number of samples. However, it requires a large memory 
to construct such a volume, which can only have a limited resolution.  
In comparison, we compute a 2D transmittance map instead of an expensive 3D 
volume, which reduces the time-complexity to be linear in the number of pixels and 
requires a much smaller amount of memory.

\section{Method}

\begin{figure*}
  \centering
  \begin{minipage}[b]{0.2500\textwidth}
      \centering
      \includegraphics[width=\textwidth]{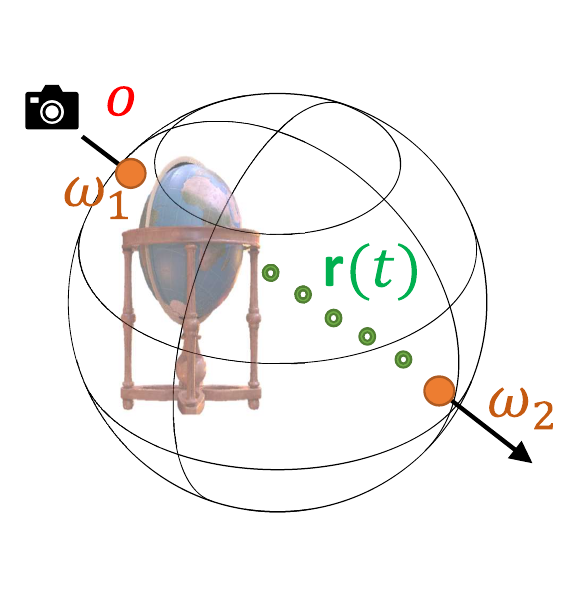}\\
  \end{minipage}
  \begin{minipage}[b]{0.300\textwidth}
      \centering
      \includegraphics[width=\textwidth]{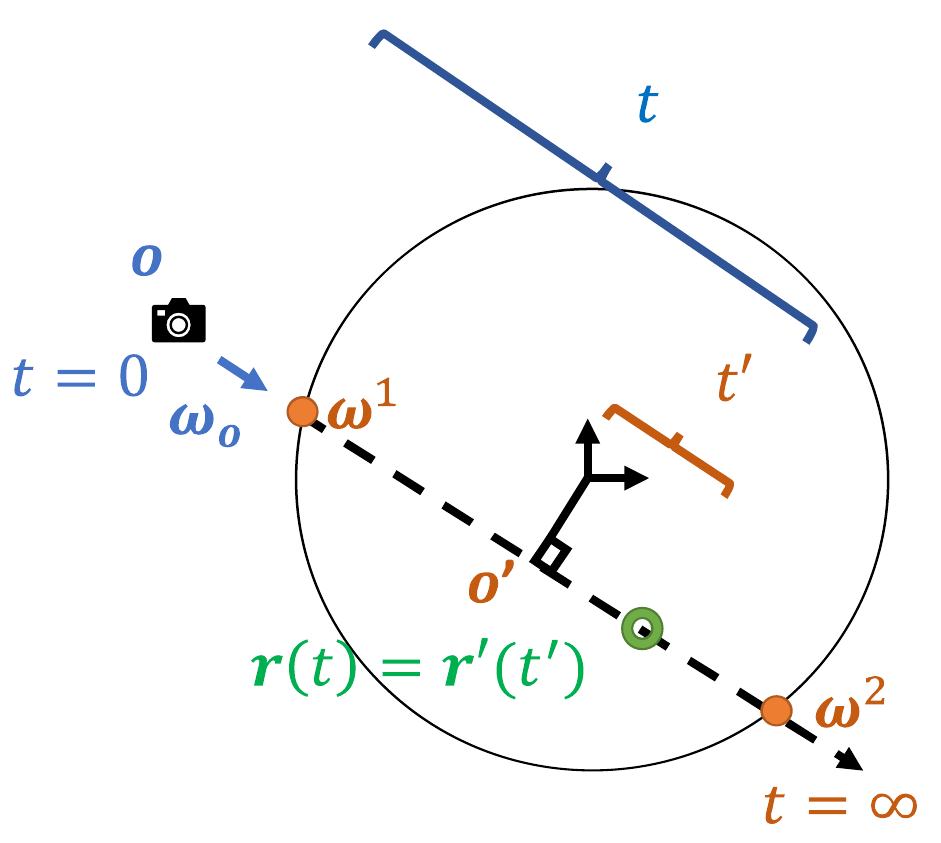}\\
  \end{minipage}
  \begin{minipage}[t]{0.35\textwidth}
    \centering
    \includegraphics[width=\textwidth]{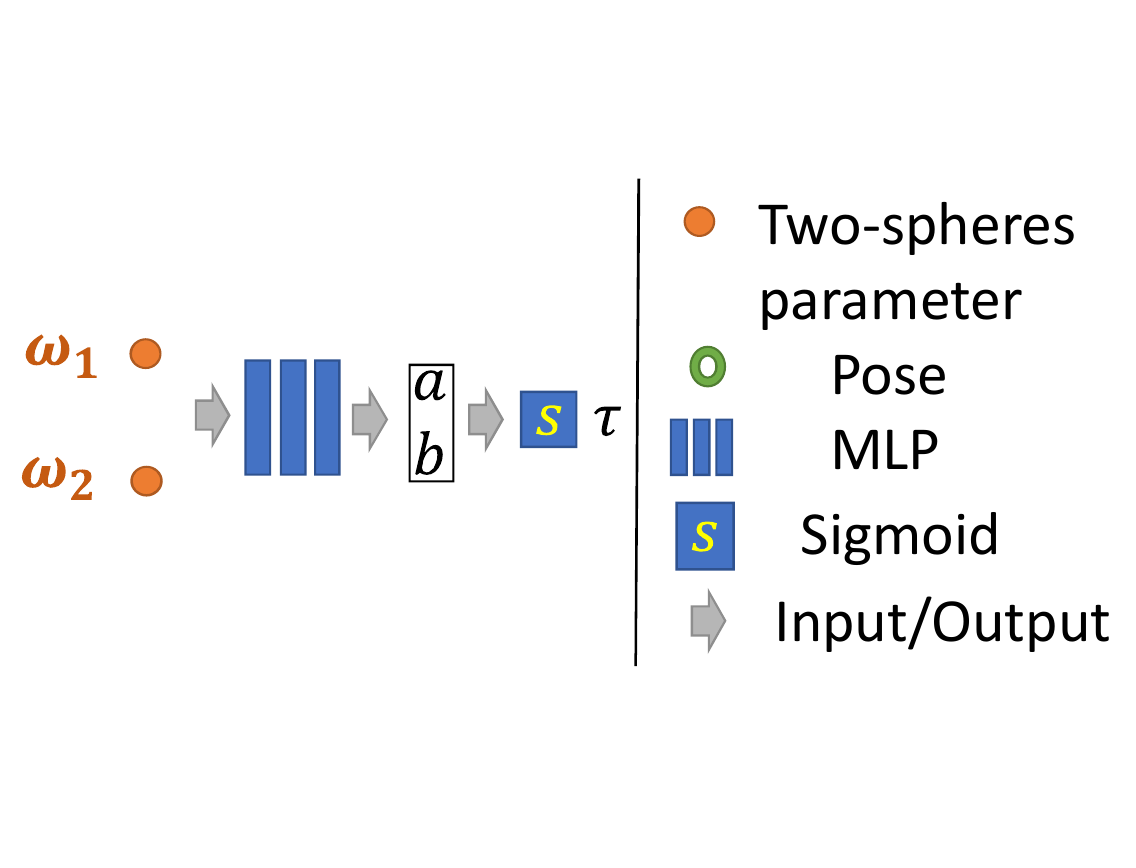} \\
  \end{minipage}
  \caption{We represent a ray by two-spheres \cite{camahort1998uniformly} representation
  i.e. two intersection points on the unit sphere (left).
  Distance to ray origin is invariant to the camera location (middle). Two-spheres
  representation is used to define the neural transmittance function (right).
    }
\label{fig:method}
\end{figure*}


  We accelerate the rendering of neural volumetric
  representations such as NRF~\cite{bi2020neural}.
  Since the dominating factor in running time is the evaluation
  of light transmittance, we train a network to predict that.
  Unlike prior work that predicts the transmittance per point 
 \cite{nerv2020}, we predict it per ray. We model it by a
  logistic function and achieve almost two orders 
  of magnitude speedup compared to baseline methods while maintaining 
  high-quality renderings.

\textbf{Ray parametrization.}
\label{sec:ray_parametrization}
We use two-spheres \cite{camahort1998uniformly} parametrization to represent a
ray independent of the camera and light source positions. Specifically,
we assume that the object of interest resides in the unit sphere centered at the world origin.
Therefore, every
ray passing through the object has two intersection
points with the unit sphere, denoted $\bm{\omega}^1$ and $\bm{\omega}^{2}$.
We use the two points as our ray representation. To sample a point $\bm{r}'(t')$ on the ray,
we define the original point $\bm{o}'$ to be the midpoint between $\bm{\omega}^{1}$ and
$\bm{\omega}^{2}$. Then we have,
\begin{equation}
\bm{r'}(t') = \bm{o}' + \bm{\omega}' t' 
\label{eq:ray_distance}
\end{equation}
where $\bm{\omega}'$ is the unit vector pointing from $\bm{\omega}^{1}$ to $\bm{\omega}^{2}$.
Two-spheres parameterization is visualized in Figure \ref{fig:method}. Note
that we use notation $'$ to distinguish the two-spheres parameterization.

\textbf{Neural Transmittance.}
\label{sec:transmittance}
NRF \cite{bi2020neural} computes the transmittance through ray marching,
by sampling $\sigma$ along the ray and querying a MLP.
This is computationally expensive, especially
for the environment map rendering where we need to compute the
incoming radiance from many directions. We train another MLP to predict
transmittance along a ray without ray marching. Hence, we achieve 100 times
speedup compared to prior work. 

The key observation is that for opaque objects, the numerically
computed transmittance can be well modeled by a sigmoid function $\bm{S}$,
(demonstrated in the supplementary material). Therefore, we train
a MLP $\bm{F}$ to predict the slope $a$ and center $b$ of a sigmoid function. 
The transmittance of a
ray $\{\bm{\omega}^1, \bm{\omega}^2\}$ can be written as,
\begin{eqnarray}
a, b &=& \bm{F(\omega^{1}, \omega^{2})} \qquad a > 0,~b\in[-1, 1] \\
\tau(\bm{r}'(t'), \omega') &=& \bm{S}(a(t'-b))
\label{eq:neural_transmittance_sigmoid}
\end{eqnarray}
The predicted transmittance $\tau$ can then be substituted in differentiable rendering equation
\eqref{eq:nerf} and \eqref{eq:nrf}. Therefore, we can train the MLP of $\bm{F}$
end-to-end, as described in the next section.

\begin{figure}
  \centering
  \begin{minipage}[b]{0.40\textwidth}
    \centering
    \includegraphics[width=\textwidth]{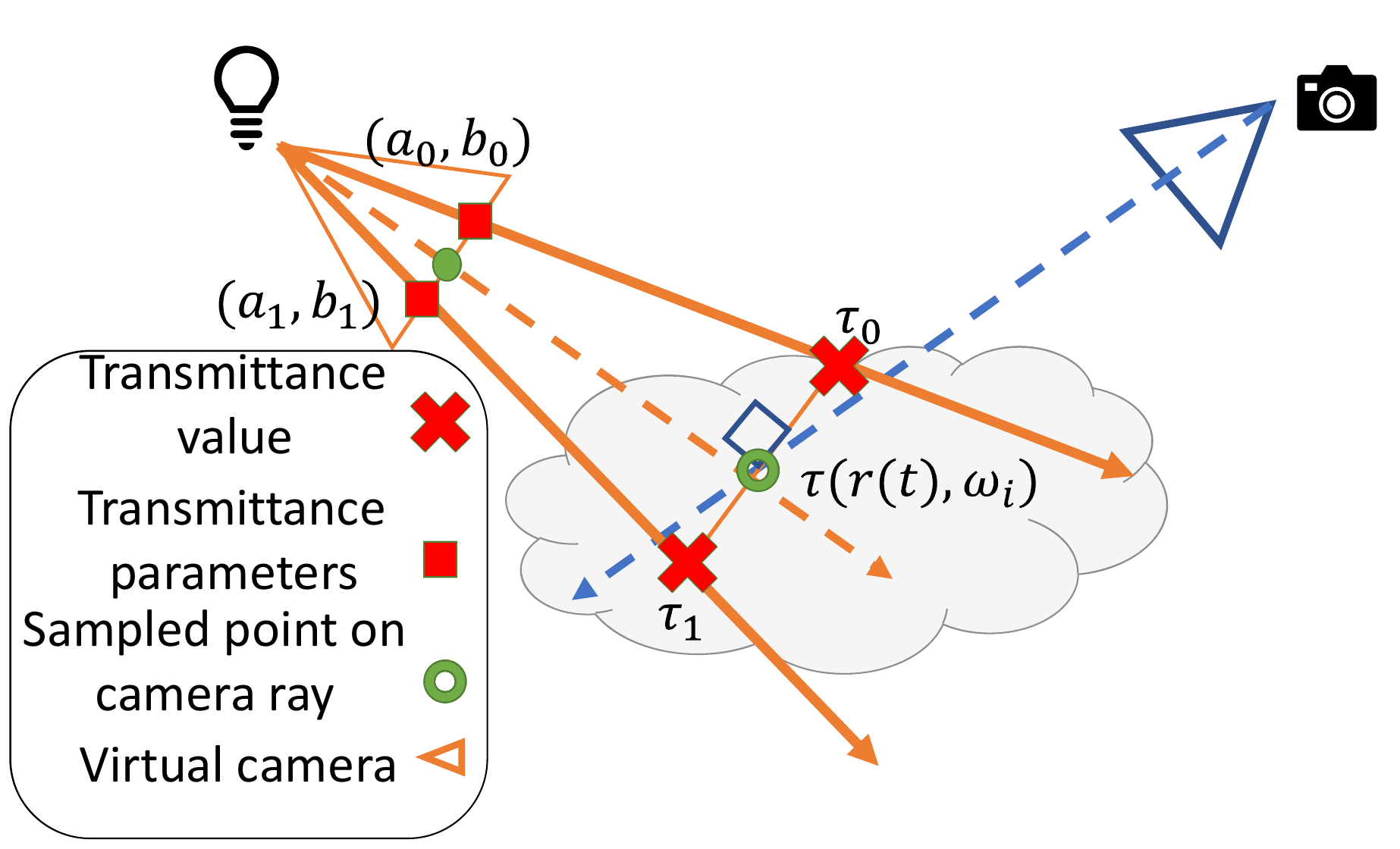} \\
\end{minipage}
\begin{minipage}[b]{0.40\textwidth}
    \centering
    \includegraphics[width=\textwidth]{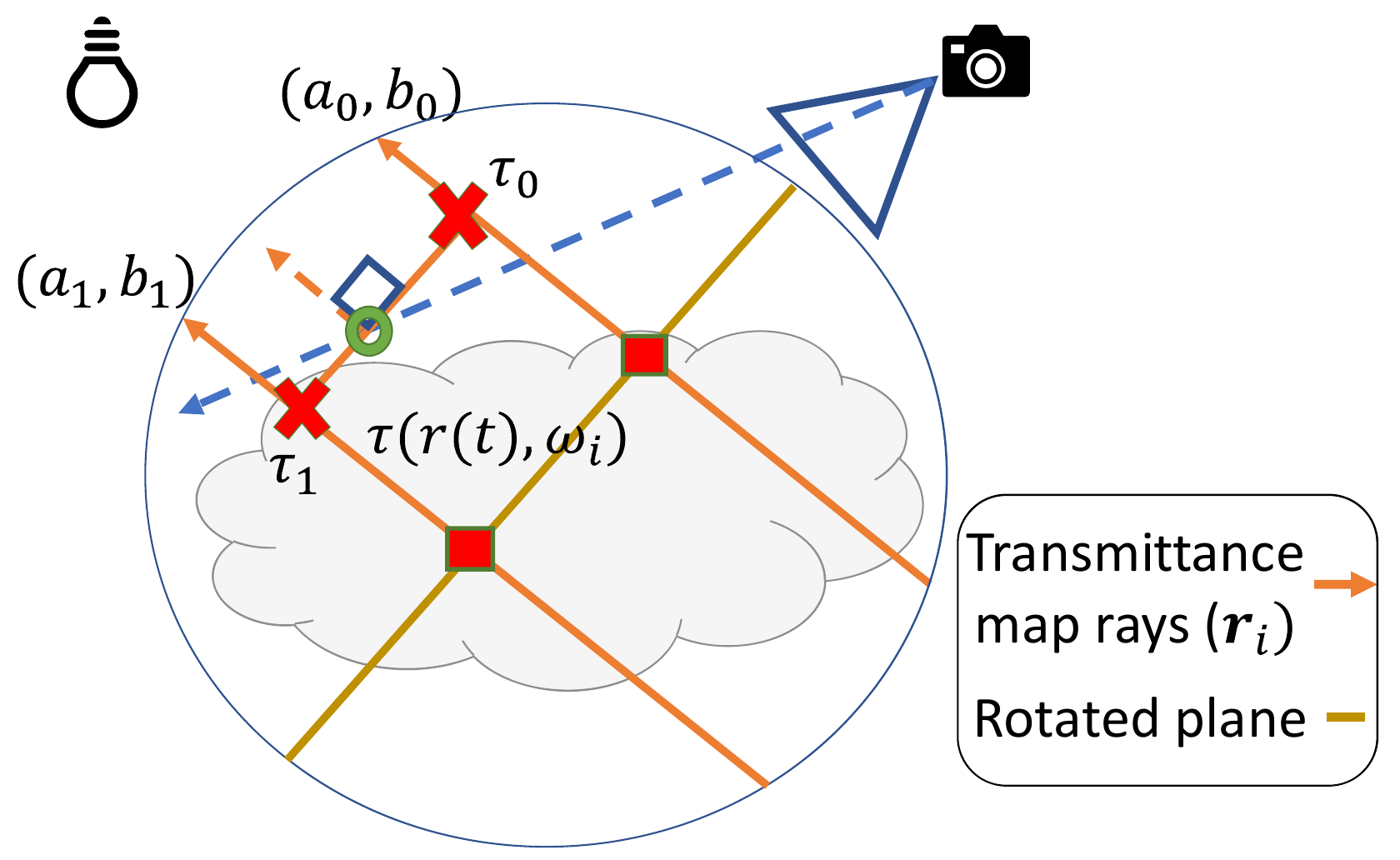} \\
    \end{minipage}
    
    \caption{Transmittance map for a point light (left) and a directional light (right).
  For each point (green dot) we first find the nearest rays on the transmittance map and
  then evaluate the neural transmittance function for interpolation (Section~\ref{sec:transmittance_map}).}
\label{fig:transmittance_map}
\end{figure}

\subsection{Training}
\label{sec:training}

Similar to Bi et al.~\cite{bi2020neural}, we train our networks on
images captured under \rebuttal{previously setup} camera and light. 
We jointly train the networks for NRF and those for transmittance. 
Given a pixel $\bm{p}$,  our loss function is as follows:
\begin{align}
 L & =  \alpha_1 L_{\text{nrf}} + \alpha_2 L_{\text{nt}} \\
  L_{\text{nrf}} & = \sum_{\bm{p}} || \bm{I}^{\bm{p}}_{\text{coarse}} - \bm{I}^{\bm{p}}||_2^2 + || \bm{I}^{\bm{p}}_{\text{fine}} - \bm{I}^{\bm{p}}||_2^2 \label{equ:loss_nrf} \\ 
  L_{\text{nt}} &= \sum_{\bm{p}}\sum_{\bm{x} \in \bm{r}(\bm{p})} || \tau_{\text{nrf}}^{\bm{x}} - \tau_{\text{nt}}^{\bm{x}} ||_2^2 +   \sum_{p} || \bm{I}^{\bm{p}}_\text{nt} - \bm{I}^{\bm{p}} ||_2^2 \label{equ:loss_nt}
\end{align}
where $\alpha_1$ and $\alpha_2$ are the weight coefficients.
Here the losses in Equation~\ref{equ:loss_nrf} minimize the difference between the
coarse prediction color 
$\bm{I}^{\bm{p}}_\text{coarse}$, fine prediction color $\bm{I}^{\bm{p}}_{\text{fine}}$
and ground truth color $\bm{I}^{\bm{p}}$. The predicted colors are calculated 
via differentiable ray marching as specified in Equation~\ref{eq:nerf} and Equation~\ref{eq:nrf}. 
The loss function $L_{\text{nrf}}$ is the same used by Bi et al. \cite{bi2020neural}.

Transmittance network loss $L_{\text{nt}}$ consists of two terms,
namely the transmittance and the color loss. 
In terms of transmittance loss, for each fine sample $\bm{x}$ on the ray $\bm{r}(\bm{p})$
corresponding to a pixel $\bm{p}$, we predict its transmittance  $\tau_{\text{nt}}^x$
using our transmittance network, and minimize its difference from the transmittance
$\tau_{\text{nrf}}^x$ calculated from the volume density predicted by the fine NRF
network (Equation~\ref{eq:nerf}). For the color loss, we calculate the pixel colors
using our predicted transmittance and minimize their difference from the ground truth colors.

\textbf{Training augmentation.}
Training the transmittance networks on the training images
fails to generalize to unseen rays, since they cover a small portion of
rays that pass the unit sphere.
Thus, we increase the number of training rays particularly for those that are
not included in the existing camera rays.
Instead of capturing more input images and increasing acquisition complexity, 
we randomly sample rays in space and use the NRF network to  
predict their color and the transmittance for points on them as 
additional supervision to our transmittance network.  
To this end we randomly sample pairs of points $\bm{\omega}_1$ and $\bm{\omega}_2$
on the sphere at each iteration, which uniquely determine a set of rays. 
These sampled rays are used to train the transmittance network only, and  
we use the same loss function as in Equation~\ref{equ:loss_nt} to train
the transmittance network except that the predicted color by the NRF network
is used for the ground truth. In Figure~\ref{fig:comparison} and the supplementary
material we visually compare relighting with and without augmentation

\subsection{Efficient rendering with precomputed transmittance map}
\label{sec:transmittance_map}

At test time, computing the transmittance of a point is computationally expensive.
Na\"ively rendering a scene with an environment map requires $l \cdot m \cdot n^2$ queries
of the transmittance network where $m,l$ and $n$ are respectively the number of pixels,
light sources and samples for ray marching. State of the art methods reduce this
time-complexity to  $m \cdot l \cdot n$~\cite{bi2020neural, nerv2020}.

We propose a novel method to accelerate the rendering in three steps.
First, we sample $m$ rays $\bm{r}_i, i \in \{0,1...,m\}$ and precompute the
transmittance parameters for each. These parameters are denoted as $a_i$ and $b_i$
and require $m \cdot l$ neural network queries.
Second, for each point of interest, we find points that lie on four closest $\bm{r}_i$.
Third, we interpolate the transmittance for the point of interest.
Algorithms are provided in the supplementary material.

\textbf{Transmittance map} is computed in two steps. We compute the rays $\bm{r}_i$
depending on the type of light source and then we compute the parameters for each ray.
For a point light, we place a virtual image plane there with $m$ pixels. $\bm{r}_i$
is the outgoing ray from pixel $i$ of the virtual image plane. For a directional light,
we sample a set of parallel rays toward the direction of the light that pass from the unit sphere.
To this end, we uniformly sample $m$ points on a circle perpendicular to the light direction
passing the world origin. $\bm{r}_i$ in this case originates on the sample point
on the circle and points toward the light source direction.

After computing rays $\bm{r}_i$, we find the two-spheres parameters by intersecting
them with the unit sphere (Section~\ref{sec:ray_parametrization}). We use our transmittance
network to precompute $a_i$ and $b_i$.
Transmittance map is depicted in Figure~\ref{fig:transmittance_map} on a 2D example.

\textbf{Nearest Points.}
During rendering, given a point of interest $\bm{x}$ and light direction $\bm{\omega}_i$,
we approximate the transmittance as
follows. We first find a set of four
rays close to $\bm{x}$ denoted as $\bm{Q}=\{\bm{r}_0,\bm{r}_1,\bm{r}_2,\bm{r}_3\}$.
We can then find a set of four close points to $\bm{x}$, namely $\bm{\mathcal{X}}=\{\bm{x}_0,\bm{x}_1,\bm{x}_2,\bm{x}_3\}$,
that reside on these rays. Using Equation~\ref{eq:neural_transmittance_sigmoid}
we can compute the transmittance values for points in $\bm{\mathcal{X}}$. Then we
approximate the transmittance on $\bm{x}$ by interpolation. These steps
are explained below and depicted in Figure~\ref{fig:transmittance_map}.

For the case of a point light, we project $\bm{x}$ on the virtual image
plane and locate its four nearest neighbor pixels and their corresponding rays
$\bm{Q}$. Four closest points in $\bm{\mathcal{X}}$ are merely the intersection
of the perpendicular plane to $\bm{\omega}_i$ that passes through $\bm{x}$ and
the nearest rays $\bm{Q}$. For rendering with a directional light, we find closest
rays $\bm{Q}$ by projecting the point $\bm{x}$ on a plane orthogonal to $\bm{\omega}_i$
that passes the origin.
We then find $\bm{\mathcal{X}}$ by intersecting the plane orthogonal to the light
direction which passes through $\bm{x}$ with four rays in $\bm{Q}$.

After finding the set $\bm{\mathcal{X}}$ of four closest points, we can compute their
transmittance by Equation~\ref{eq:neural_transmittance_sigmoid} and find the 
transmittance on $\bm{x}$ using bilinear interpolation. As shown in Figure~\ref{fig:comparison}
and Table~\ref{tbl:time_ours}, our method achieves almost two orders of magnitude speedup
compared to baseline methods without sacrificing the image quality.

\textbf{Implementation.}
\label{sec:implementation}
We follow the training protocols of NRF. The input of our
algorithm is a set of images taken from an object with a mobile phone with flashlight
in a darkroom. We capture around 400 per object. The camera poses and intrinsic
parameters are estimated with COLMAP~\cite{schoenberger2016sfm}. 
We translate and scale the scene to fit the reconstructed geometry into a unit sphere at the origin.
We jointly train the neural networks for NRF and the transmittance network 
with a batch size of $4$.
For each batch, we randomly select a training image and sample $16 \times 16$
pixels for training. In addition, we also randomly sample $128$ rays 
as described in Section~\ref{sec:training} to generate augmented training 
data for the transmittance networks. We optimize the networks using 
Adam optimizer~\cite{AdamSolver} with a learning rate of $2e^{-5}$.
The networks are trained on $4$ Nvidia RTX 2080 GPUs, and the full training 
takes around 2 days.

\vspace*{-.15in}
\section{Results}
\label{sec:results}
\begin{table}
    \small
    \centering
\sisetup{detect-weight,mode=text}
\renewrobustcmd{\bfseries}{\fontseries{b}\selectfont}
\newrobustcmd{\B}{\bfseries}
\small
\centering
\begin{tabular}{|c|S[table-format=4.2]|S[table-format=4.2]|S[table-format=4.2]|S[table-format=4.2]|S[table-format=4.2]|S[table-format=4.2]|} \hline
  \multirow{2}{*}{Scene} & \multicolumn{3}{c|}{Transmittance} & \multicolumn{3}{c|}{Overall} \\ \cline{2-7}
                  & \text{Ours}     & \text{NRF}      & \text{Speed up} & \text{Ours}     & \text{NRF}      & \text{Speed up}   \\ \hline 
  Point light     & 0.25            & 96.73           & \B 384.10    & 65.49          & 162.88           & \B 2.48 \\ \hline 
  Environment map & 122.38           & 47806.81         & \B 390.62  & 522.45          & 48151.80        & \B 92.16 \\ \hline 
    \end{tabular}
    \caption{Runtime of our method compared to NRF in seconds. Our method is
    \speedupPointlight~times faster than NRF for relighting with a
    point light and \speedupFiveHunred~times faster for relighting
    with a $50 \times 10$ environment map.
    }
    \label{tbl:time_ours}
\end{table}

We evaluate our method in three aspects. First, we demonstrate
that our method is more than \speedupFiveHunred~times faster than Bi et al.\cite{bi2020neural}.
Second, we show that our augmentation method is necessary for neural transmittance to avoid
overfitting. Third, we show that, even though our method includes approximation
steps, our results are still qualitatively and quantitatively comparable to prior work.
We run our experiments on 3 synthetic scenes, namely,
\textit{Happy Buddha}, \textit{Sitting Buddha} and \textit{Globe} and 2 real scenes, namely,
\textit{Girl} and \textit{Pony}. All of the synthetic scenes, Girl and Pony are
rendered with 600$\times$600, 592$\times$478 and 763$\times$544 resolution images respectively.
We use 500 images to train the synthetic scenes and respectively
384 and 380 images for training the Girl and Pony. We show our results on view synthesis
and relighting with point light on an image sequence in the supplementary material.

\begin{figure}
  \centering
  \begin{minipage}[b]{0.70\textwidth}
  \includegraphics[width=\textwidth]{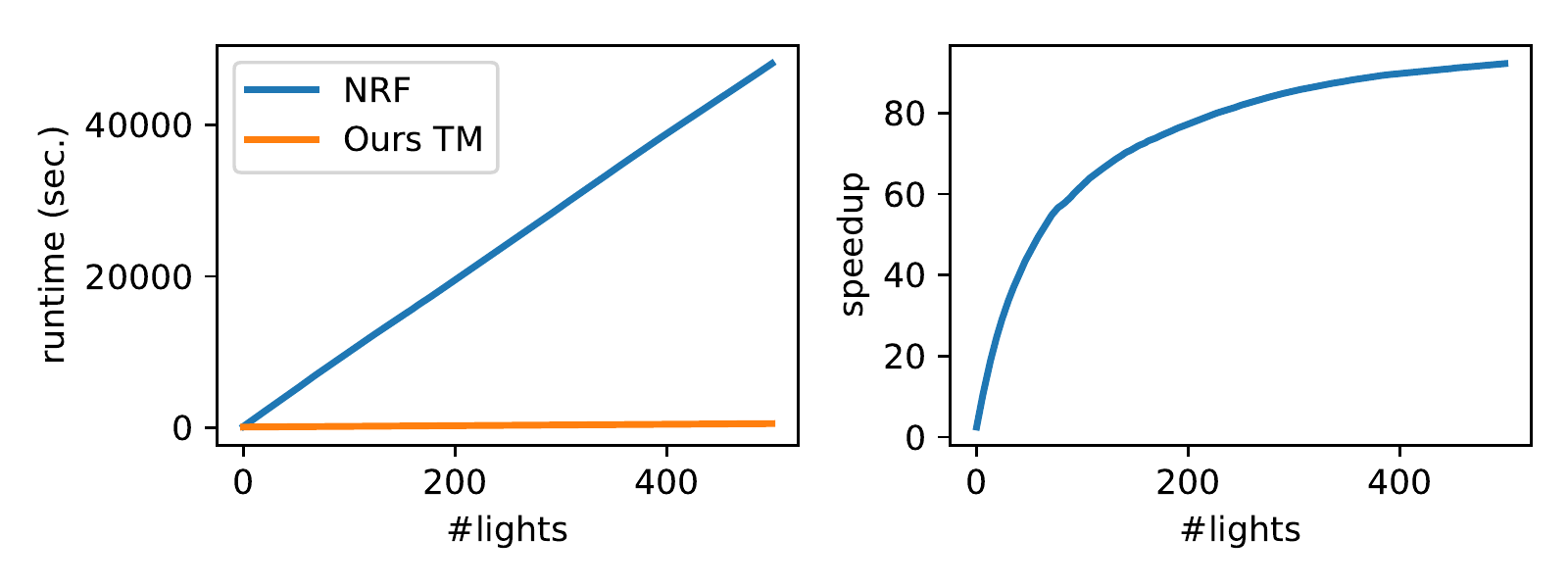}
  \end{minipage}
\caption{Overall runtime for relighting of the Happy Buddha scene (left). We
relight the scene with many environment maps each with a unique resolution.
The speed up is the division of NRF runtime by our method with Transmittance Map (TM),
shown on the right.}
\label{fig:runtime_scenes}
\end{figure}

\begin{figure}[!htb]
    \centering
    \begin{minipage}[b]{0.80\textwidth}
      \includegraphics[width=\textwidth]{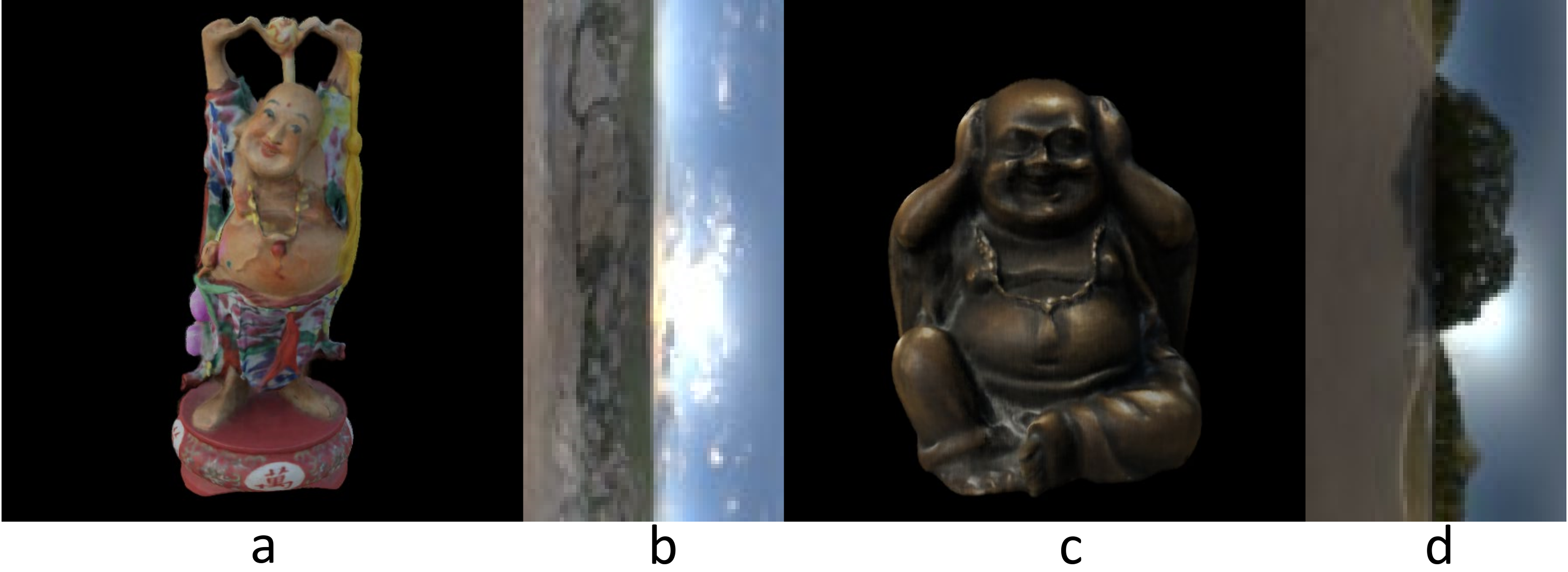}
    \end{minipage}
       \caption{We render Happy Buddha (a) and Sitting Buddha (b) with $64 \times 128$ environment maps
       (b and d). It takes 2.7 hours with our method to render each of these scenes and 
       potentially more than 251 hours with prior work.}
  \label{fig:large_envmap}
  \end{figure}

\textbf{Runtime.} We compare our method with that of Bi et al. \cite{bi2020neural}.
We gain more (\speedupFiveHunred~times) speed up for rendering a scene under environment maps
with higher resolution than those with low resolution (\speedupPointlight~times).
We plot the runtime of our method and Bi et al. \cite{bi2020neural} in Figure~\ref{fig:runtime_scenes}.
To this end, we run both methods on 500 different environment maps. Each of these
maps have a unique resolution, namely, $i \in \{1, 2, 3, ..., 500\}$ pixels.
We compute the runtime for these experiments and the results are shown in Figure~\ref{fig:runtime_scenes}.
Figure~\ref{fig:runtime_scenes} shows the runtime of both methods and the speed up
of our method compared to Bi et al. \cite{bi2020neural}. \rebuttal{We show that our method allows
rendering with high resolution environment map in Figure \ref{fig:large_envmap}.
NRF \cite{bi2020neural} takes more than 251 hours to render in this resolution.}

\begin{figure}[!htb]
  \centering
  \begin{minipage}[b]{0.80\textwidth}
    \includegraphics[width=\textwidth]{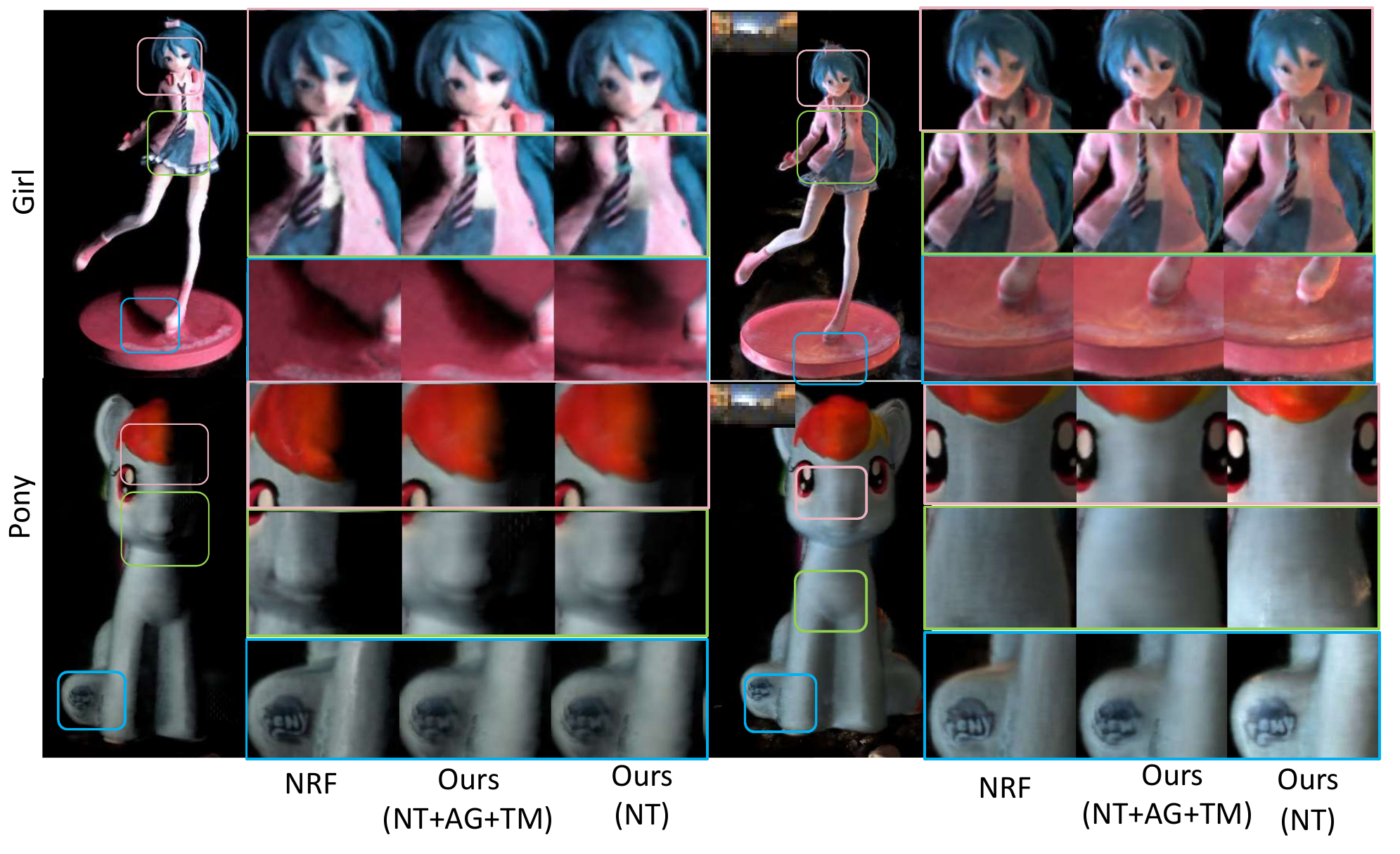}
  \end{minipage}
  \begin{minipage}[b]{0.80\textwidth}
    \includegraphics[width=\textwidth]{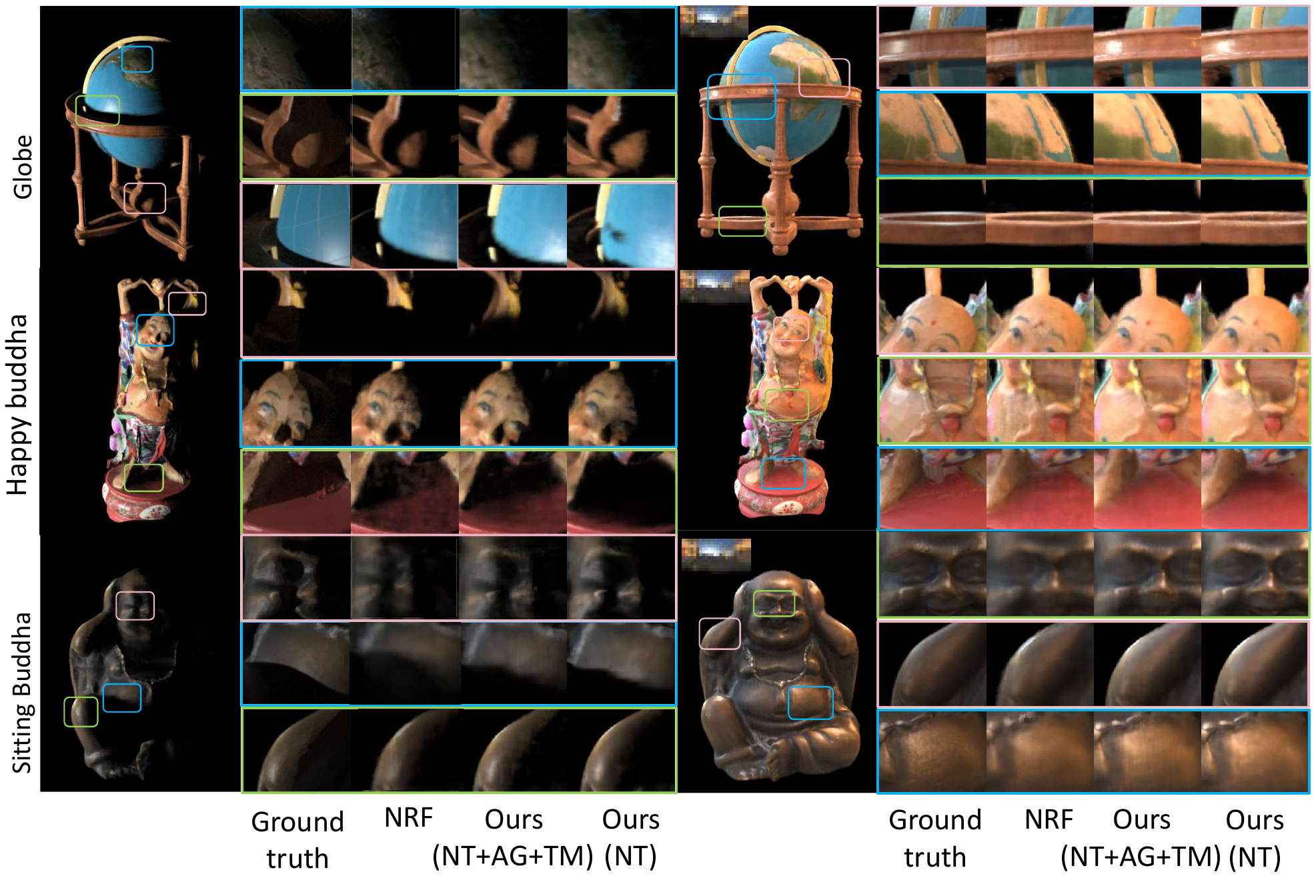}
  \end{minipage}
     \caption{ Relighting results with a point light (left) and
     with $24 \times 12$ pixels environment map (right). The quality of relighting with our method
     is comparable to that of NRF and ground truth. We compare our method with Neural Transmittance (NT)
     only and with augmentation (AG) and Transmittance Map (TM). Augmentation removes the overfitting artifacts.}
\label{fig:comparison}
\end{figure}

The speed up curve in Figure~\ref{fig:runtime_scenes} shows that our method is faster for larger environment
maps compared to smaller ones. In more detail, the bottle neck for the runtime of neural rendering is neural network
queries. These queries are required to retrieve reflectance and transmittance values.
Our method, similar to NRF, requires $n \cdot m$ queries for reflectance.
Additionally, our method requires $l \cdot m$ queries for transmittance while NRF requires
$l \cdot m \cdot n$. As a result, the ratio of overall queries for NRF compared
to ours is $\psi = \frac{v_{NRF}}{v_{ours}}=\frac{l \cdot m \cdot n + n \cdot m}{l \cdot m + n \cdot m}$
where $v_{NRF}, v_{ours}$ and $\psi$ are respectively the time complexity of NRF~\cite{bi2020neural},
that of our method and the hypothetical speed up of our method. For one light source this is a constant $\frac{2n}{n+1}$
and for large number of light sources it approaches to $n$ that is the sample count along a ray.
Sample count is normally chosen to be 192 which suggests that our method should be almost 192 times
faster than Bi et al.\cite{bi2020neural} for large environment maps.
Our speed up plot in Figure~\ref{fig:runtime_scenes} however, converges to a constant
value around \speedupFiveHunred. This value is lower than the hypothetical speed up (192) since our
algorithm requires additional steps including nearest neighbor lookup and interpolation. It is also clear from Table~\ref{tbl:time_ours}
which shows that our method is \speedupPointlight~$\times$ faster for relighting with uncollocated point light.
Table~\ref{tbl:time_ours} moreover shows that our method is more than
\speedupTransmittance $\times$ faster for in the precomputation stage compared to NRF.
In the supplementary material we compare the time for creating transmittance map
compared to transmittance volume of Bi et al~\cite{bi2020neural} for different resolutions.

\textbf{Parametrization and Augmentation.} The two-spheres parametrization enables
us to use a monotonic function for the definition of neural transmittance. Naively fitting such a
function only over the view directions of input images would lead to
overfitting and our augmentation method leads to higher accuracy. To
demonstrate this, we compute the accuracy of our method with and without augmentation
in Table \ref{tbl:comparison_accuracy} for both cases of relighting the synthetic scenes
with point light and environment map. For most of the cases the
augmentation method leads to higher accuracy by either or both metrics.
We visually show the necessity of augmentation in Figure~\ref{fig:comparison} and in the
supplementary material. In Figure~\ref{fig:comparison} we compare our method with Neural
Transmittance (NT) only and to that with augmentation and transmittance map. This figure
shows that the overfitting artifacts are removed by our augmentation method. We show an
ablation study in the supplementary material.

\textbf{Accuracy.}
\label{sec:accuracy}
Although our method uses many approximation steps in favor of efficient rendering,
it is still quantitatively and qualitatively comparable to 
prior work. We show this in Table~\ref{tbl:comparison_accuracy} 
which includes the error values of different steps of our algorithm. These errors are computed
on synthetic scenes under point light and environment map lighting. 
We evaluate our method quantitatively on 400 and 18 images respectively
for point light and environment lighting. In the first 200, the light source
moves around the object on a sphere. In the next 200, the view angle also moves
on the same position as light. For environment map cases, 
we relight each scene with 18 different environment maps from a fixed view point.
We compare each of these images with the ground truth and take the average error.
In most of the rows in Table~\ref{tbl:comparison_accuracy}, our method is
quantitatively similar or slightly worse than NRF despite the many approximation steps.
Nevertheless, the resulting error is small.

\begin{table*}
  \small
    \centering
\renewcommand{\arraystretch}{0.9}
\setlength{\tabcolsep}{1.0pt}
\begin{tabular}{|c|cccc|cccc|}
  \hline
\multirow{2}{*}{Scene} & NRF & Ours TM & Ours AG & No AG & NRF & Ours TM & Ours AG & No AG   \\ \cline{2-9} 

& \multicolumn{8}{|c|}{Point light} \\  
\cline{2-9} 
&  \multicolumn{4}{|c|}{MS-SSIM} &  \multicolumn{4}{|c|}{LPIPS} \\ \hline  
Happy Buddha  & 0.963  & 0.954  & 0.954  & 0.948  & 0.049  & 0.055  & 0.055  & 0.058 \\ \hline  
Sitting Buddha  & 0.985  & 0.973  & 0.973  & 0.965  & 0.066  & 0.049  & 0.049  & 0.042 \\ \hline  
Globe  & 0.681  & 0.666  & 0.666  & 0.668  & 0.219  & 0.229  & 0.229  & 0.228 \\ \hline  
&  \multicolumn{4}{|c|}{rMSE} &  \multicolumn{4}{|c|}{SSIM} \\ \hline  
Happy Buddha  & 0.051  & 0.060  & 0.059  & 0.058  & 0.955  & 0.945  & 0.945  & 0.939 \\ \hline  
Sitting Buddha  & 0.011  & 0.022  & 0.022  & 0.026  & 0.976  & 0.967  & 0.967  & 0.959 \\ \hline  
Globe  & 0.107  & 0.126  & 0.126  & 0.127  & 0.739  & 0.732  & 0.732  & 0.732  \\ \specialrule{.1em}{.05em}{.05em} 
& \multicolumn{8}{|c|}{Environment map}  \\    \cline{2-9} 
&  \multicolumn{4}{|c|}{MS-SSIM} &  \multicolumn{4}{|c|}{LPIPS} \\ \hline 
Happy Buddha  & 0.958  & 0.937  & 0.938  & 0.940  & 0.071  & 0.077  & 0.077  & 0.076 \\ \hline 
Sitting Buddha  & 0.952  & 0.903  & 0.890  & 0.870  & 0.095  & 0.085  & 0.090  & 0.075 \\ \hline 
Globe  & 0.911  & 0.859  & 0.837  & 0.840  & 0.086  & 0.107  & 0.114  & 0.117  \\ \hline 
&  \multicolumn{4}{|c|}{rMSE} &  \multicolumn{4}{|c|}{SSIM} \\ \hline 
Happy Buddha  & 0.148  & 0.201  & 0.201  & 0.190  & 0.937  & 0.914  & 0.914  & 0.916 \\ \hline 
Sitting Buddha  & 0.041  & 0.081  & 0.085  & 0.097  & 0.939  & 0.895  & 0.880  & 0.864 \\ \hline 
Globe  & 0.159  & 0.240  & 0.281  & 0.281  & 0.899  & 0.851  & 0.831  & 0.835 \\
\hline 

\end{tabular}
\caption{We compare the accuracy of our method to the baseline. Our approximation method with
Transmittance Map (Ours TM), has only a minimal drop of accuracy compared to NRF. We achieve
higher accuracy by Augmentation (Ours AG) than no augmentation (No AG). 
\rebuttal{(rMSE, (MS-)SSIM and LPIPS respectively stand for Root Mean Squared Error, (Multi-Scale)
Structure Similarity Index Measure and Learned Perceptual Image Patch Similarity.)} }
  \label{tbl:comparison_accuracy}
\end{table*}

\vspace*{-.1in}
\section{Discussion}
We introduced Neural Transmittance and Transmittance Map which allow more than
\speedupFiveHunred $\times$ faster rendering with Neural Reflectance
Fields under environment maps and almost \speedupPointlight $\times$ 
under point lights. Despite the necessary approximations, our method
is qualitatively and quantitatively comparable to prior work. We also introduced an
augmentation method that avoids overfitting of transmittance. We expect to see this
augmentation method used for other quantities defined on the light field
domain such as radiance, depth, reflectance, etc.

An interesting future direction is to formulate a physically consistent transmittance
function to derive the volume density. Such a formulation would allow
training only one network for both of these quantities and potentially for all of
those required for volumetric rendering. In our method, these two quantities
are decoupled that results in inaccuracies such as intensity shift. However, this
does not lead to significant error as demonstrated in Section \ref{sec:accuracy}.
\section{Acknowledgement}
The authors thank Manmohan Chandraker for helpful discussion. This work was funded
in part by the Ronald L. Graham Chair, ONR grants N000141912293, N000141712687
and NSF grant 1730158.

\bibliography{references}
\end{document}


\maketitle
\section{Architecture.}
\begin{figure}[!htb]
  \centering
  \begin{minipage}[b]{0.45\textwidth}
    \includegraphics[width=\linewidth]{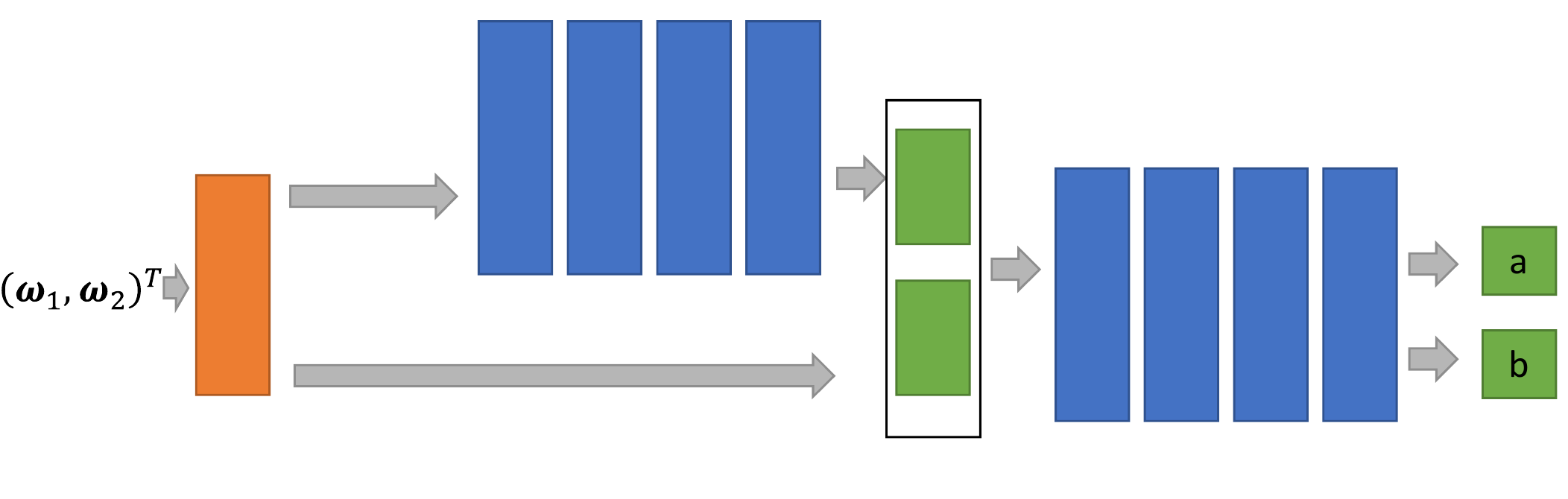}
  \end{minipage}
  \begin{minipage}[b]{0.2\textwidth}
    \includegraphics[width=\linewidth]{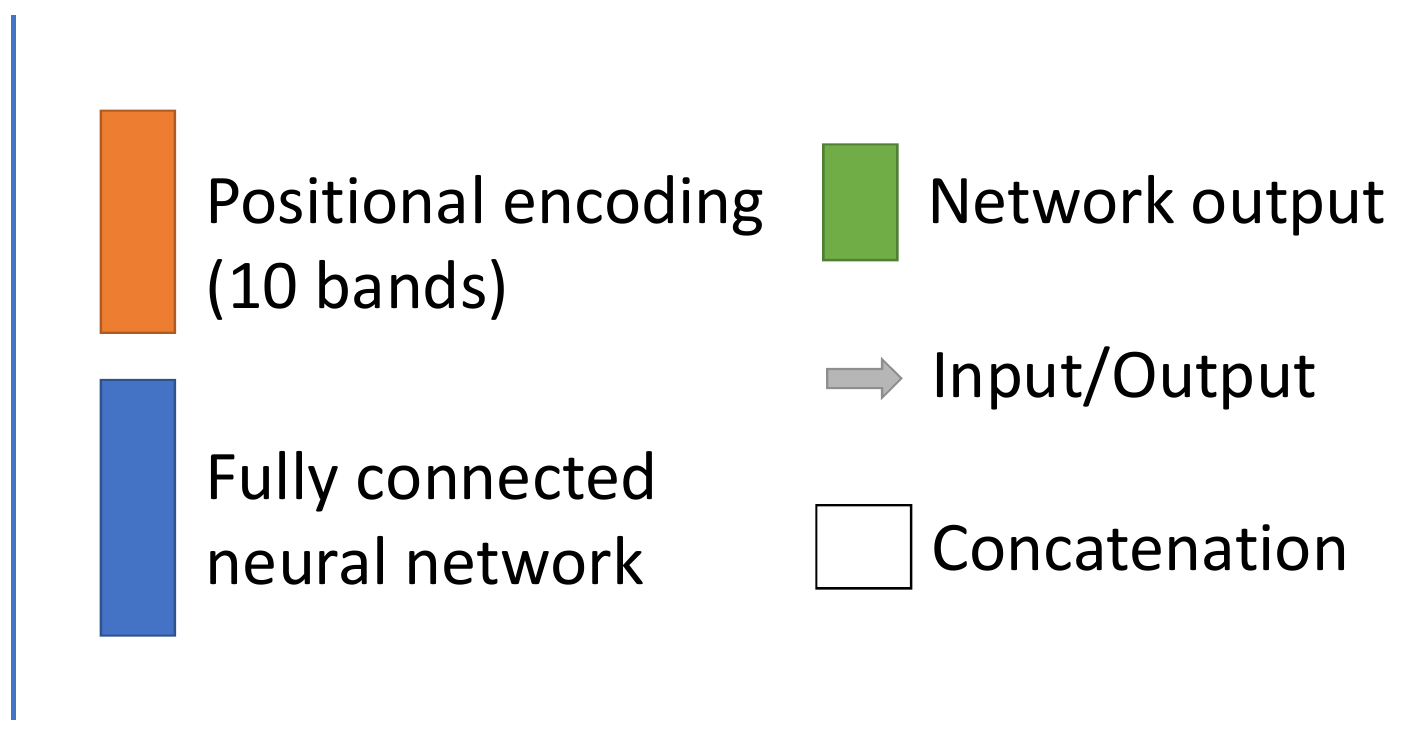}
  \end{minipage}
     \caption{We use the same neural network architecture as that of Mildenhall et al.~\cite{mildenhall2020nerf}
     and Bi et al.~\cite{bi2020neural}.}
    \label{fig:architecture}
\end{figure}

Our neural network architecture is identical to that used by Mildenhall et al.~\cite{mildenhall2020nerf}
and Bi et al.~\cite{bi2020neural}. The architecture is depicted in Figure~\ref{fig:architecture}.

\section{Transmittance map}
In the main paper, we proposed efficient rendering with precomputed transmittance map
(main paper section 4.2). Creating transmittance map is described by Algorithm~\ref{alg:transmittance_map}.
After its creation, it can be used to compute the transmittance for each point of interest $\bm{x}$
and $\bm{r}$. To this end, we first find the four closest rays to $\bm{x}$, namely
$\bm{Q}=\{\bm{r}_0,\bm{r}_1,\bm{r}_2,\bm{r}_3\}$ and the closest points to $\bm{x}$
that reside on those ray, denoted as $\bm{\mathcal{X}}=\{\bm{x}_0,\bm{x}_1,\bm{x}_2,\bm{x}_3\}$.
Then we find the transmittance value for points in $\bm{\mathcal{X}}$ as described in
Algorithm~\ref{alg:transmittance_nearest_rays}. Having the transmittance of four
closest points to $\bm{x}$, we can use interpolation to compute the transmittance
on point $\bm{x}$ as in Algorithm~\ref{alg:transmittance_interpolation}.

\begin{algorithm}[!htb]
  \SetAlgoLined
  $\bm{x}_i \gets$ $i$th pixel of a virtual image plane with $m$ pixels

  $\bm{p}_i \gets$ $i$th point of a set uniformly sampled on a circle

  \KwResult{$(a_i,b_i) \in \{1,...,m\}$}

  \ForEach(){$i \in \{1,2...,m\}$}{
    
    \uIf(){Light source is point light}{
      $\bm{q} \gets$ Compute two-spheres parameters of pixel $\bm{x}_i$ (sec. 4)
     
      $\{a_i,b_i\} \gets \{ a(\bm{q}),b(\bm{q}) \}$ (sec. 4)
    }
    \ElseIf{Light source is directional}{
            $\bm{p}'_i \gets$ Rotate $\bm{p}_i$ toward light
            
            $\bm{q} \gets$ Compute two-spheres parameters of point $\bm{p}'_i$ in direction of $\bm{\omega}_i$
            (sec. 4)
           
            $\{a_i,b_i\} \gets \{ a(\bm{q}),b(\bm{q}) \}$ (sec. 4)
            
            }
      }
  
      \caption{Transmittance map (references are to the main paper)}
      \label{alg:transmittance_map}
    \end{algorithm}

\begin{algorithm}[!htb]
  \SetAlgoLined
  \KwResult{$(\bm{q}_i,t_i) i \in \{1,2,3,4\}$}
  $\bm{\omega}_i \gets $ light direction on point $\bm{p}$

  \ForEach(){$i \in {1,2,3,4}$}{
    \uIf(){Light source is point light}{
      $\bm{x} \gets$ Project $\bm{p}$ to virtual image plane

      $\bm{y}_i \gets$ Find $i$th nearest points on image plane to $\bm{x}$
        
      $\bm{q} \gets$ Compute two-spheres parameters of pixel $\bm{y}_i$ (sec. 4)

      $\bm{q}' \gets$ Intersect the orthogonal plane to $\bm{\omega}_i$ that passes point $\bm{p}$ with $\bm{q}$

      $t_i = \norm{\bm{o}'(\bm{q}) - q'}$ (eq. 5)
      
      $\tau_i \gets \tau(\bm{q},t_i)$ (eq. 7)

    }
    \ElseIf(){Light source is directional}{
      
      $\bm{p}' \gets $ Project $\bm{p}$ on orthogonal plane to $\bm{\omega}_i$
      
      $\bm{u} \gets$ Rotate $\bm{p}'$ to $xy$ plane
      
      $\bm{v}_i \gets$ Find $i$th closest rays
      
      $\bm{q} \gets$ Intersect the orthogonal plane to $\bm{\omega}_i$ that passes $\bm{p}$ with ray $\bm{v}_i$
      
      $t_i \gets \norm{q-p'}$
      
      $\tau_i \gets \tau(\bm{q},t_i)$ 
      
    }
  }
    \caption{Nearest rays and points on transmittance map (references are to the main paper)}
    \label{alg:transmittance_nearest_rays}
\end{algorithm}

\begin{algorithm}[!htb]
  \SetAlgoLined
  \KwResult{$\tau(\bm{p},\bm{\omega}_o)$}
    $l \gets$ number of light sources

    $n \gets$ number of sampled points in space

    $\bm{P}$ set of all sampled points along view direction

    $\bm{a}, \bm{b} \gets$ initialize transmittance map alg.~\ref{alg:transmittance_map}

    \ForEach(){i $\in$ $\{1,2,...,l\}$}{

        \ForEach(){$\bm{p} \in \bm{P}$}{
        
        $\bm{Q} \gets$ four nearest rays to $\bm{p}$ towards $i$th light source alg.~\ref{alg:transmittance_nearest_rays}
        
        //Compute transmittance on neighboring rays

        \ForEach(){$\bm{q} \in \bm{Q}$}{
          
          
          
          $q' \gets$ find a point on ray $\bm{q}$ close to $\bm{r}(t)$ alg.~\ref{alg:transmittance_nearest_rays}
          
          $t_k \gets$ distance of $\bm{q}'$ to ray origin (eq. 5)
          
          $\tau_k, \gets \tau(\bm{q},t_k)$ (eq. 7)
          }
          
          //Compute $\tau(\bm{p},\bm{\omega}_o)$ by bilinear interpolation

        $\bm{u} = \frac{\bm{p}_0 - \bm{p}_1}{\norm{\bm{p}_0 - \bm{p}_1}} $

        $\bm{v} = \frac{\bm{p}_0 - \bm{p}_2}{\norm{\bm{p}_0 - \bm{p}_2}} $
        
        $\bm{\alpha} = \langle \bm{p} - \bm{q}_0, \bm{u} \rangle$
        
        $\bm{\beta} = \langle \bm{p} - \bm{q}_0, \bm{v} \rangle$

        $\tau_{t} \gets \bm{\alpha}_x \tau_0 + (1-\bm{\alpha}_x) \tau_1$
        
        $\tau_{b} \gets \bm{\alpha}_x \tau_2 + (1-\bm{\alpha}_x) \tau_3$

        $\tau(\bm{p},\bm{\omega}_o) \gets \bm{\beta}_y \tau_{t} + (1-\bm{\beta}_y) \tau_{b}$

      }
    }
    \caption{Transmittance map and interpolation step (equation references are to the main paper)}
    \label{alg:transmittance_interpolation}
\end{algorithm}

\section{Additional Results}
\begin{figure*}
  \centering
  \begin{minipage}[b]{0.42\textwidth}
    \centering
    \includegraphics[width=\textwidth]{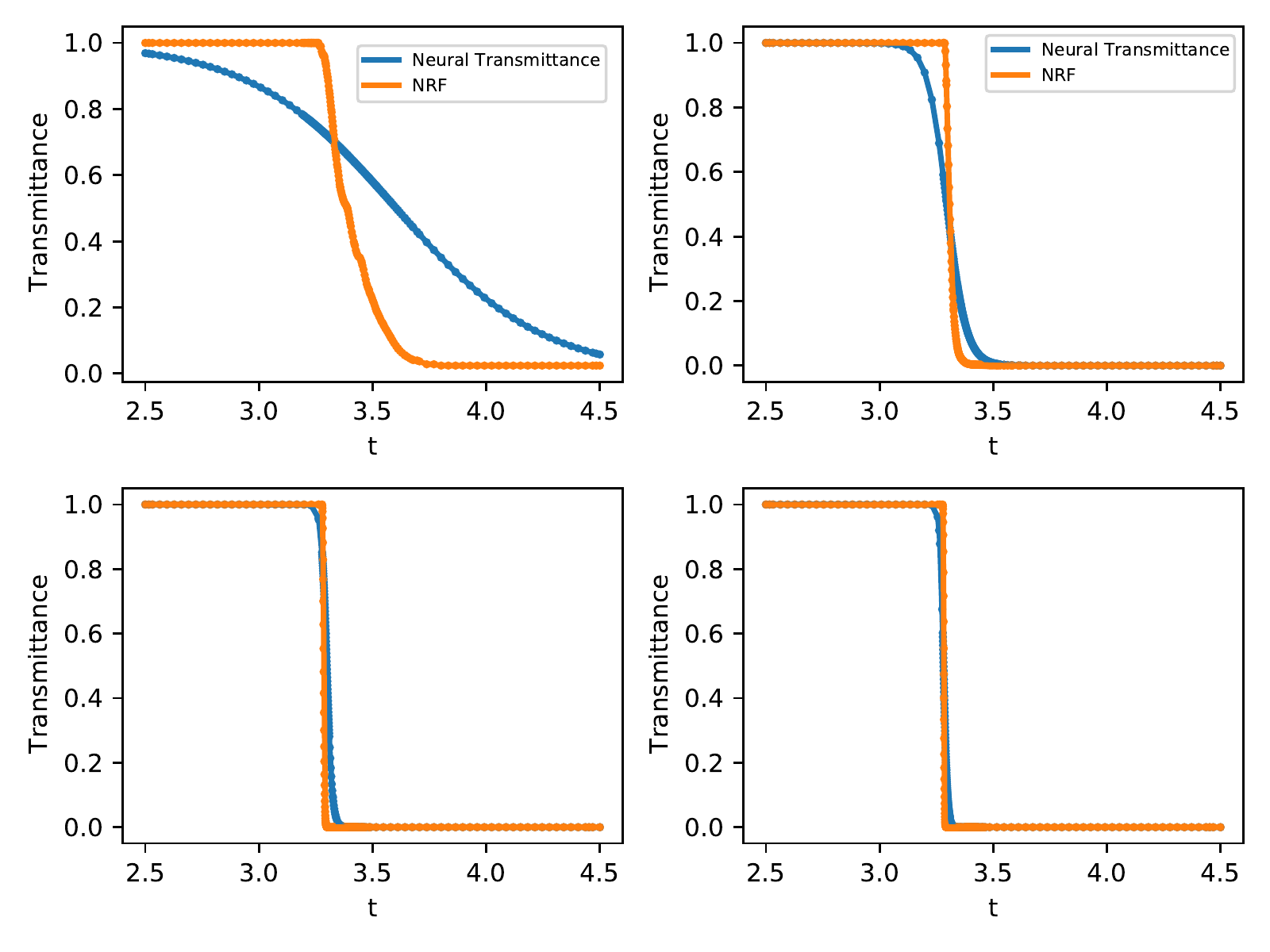}\\
    (a)
\end{minipage}
\begin{minipage}[b]{0.42\textwidth}
  \centering
  \includegraphics[width=\textwidth]{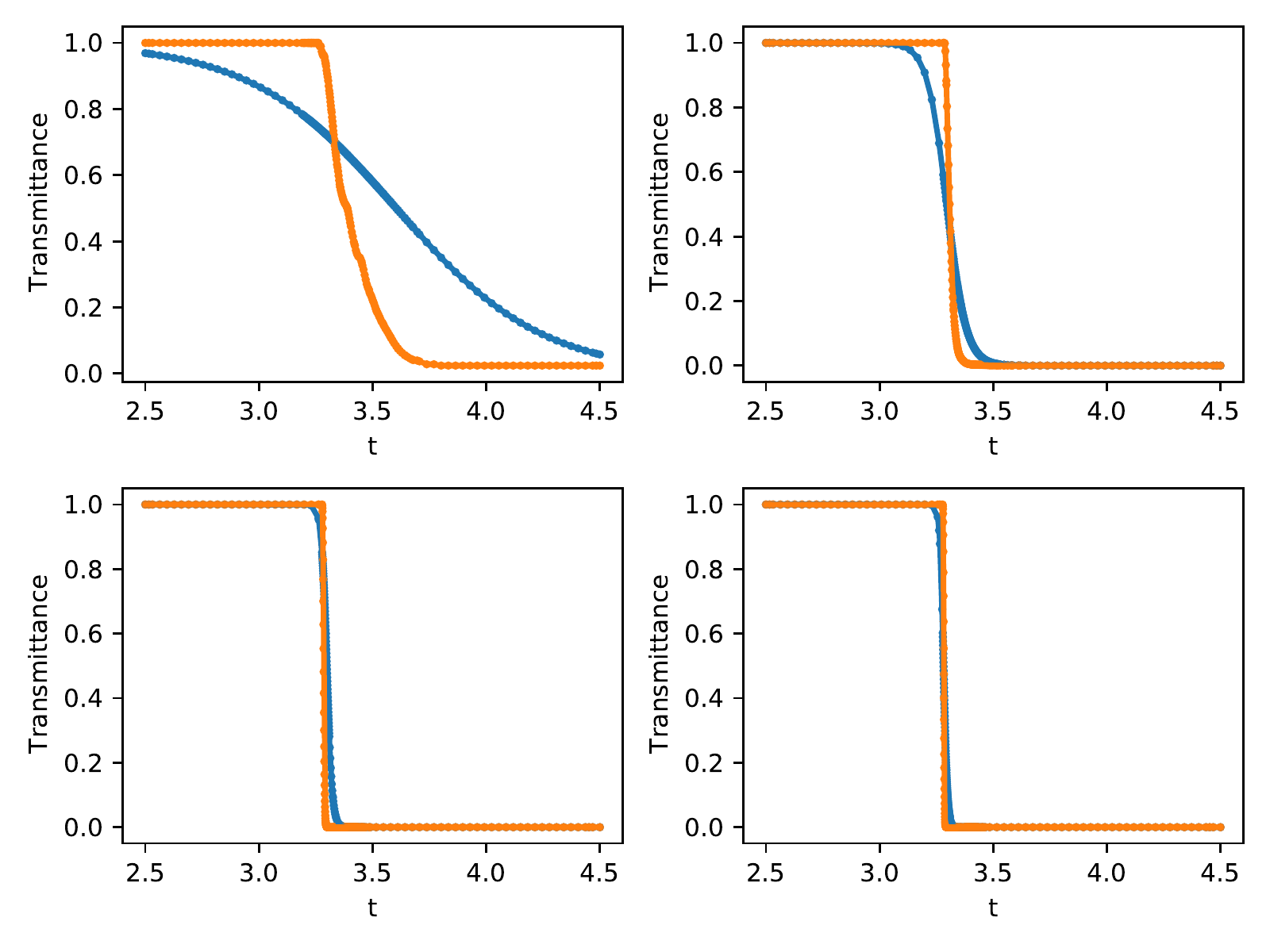}\\
  (b)
\end{minipage}

\begin{minipage}[b]{0.42\textwidth}
  \centering
  \includegraphics[width=\textwidth]{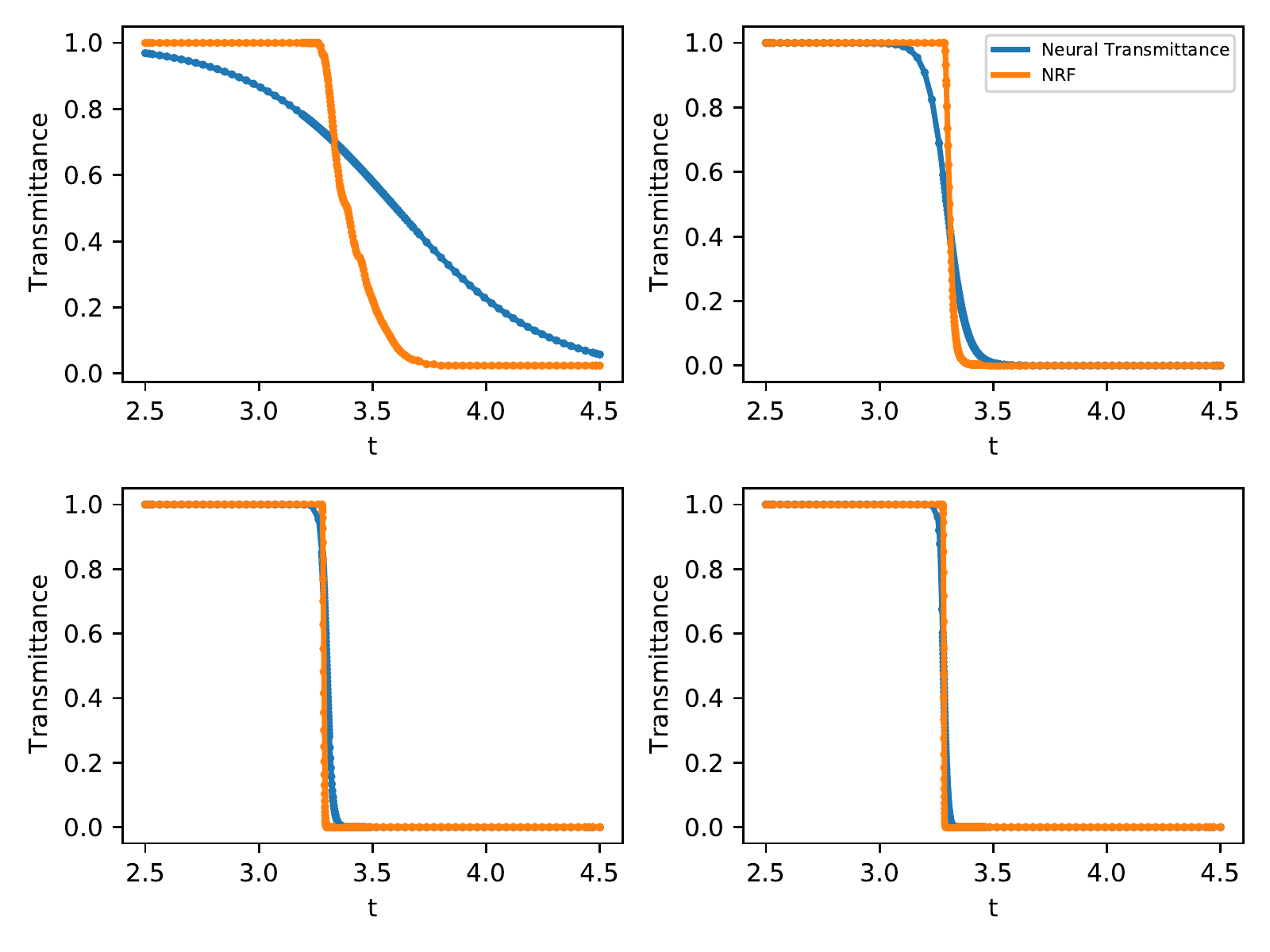}\\
  (c)
\end{minipage}
\begin{minipage}[b]{0.42\textwidth}
  \centering
  \includegraphics[width=\textwidth]{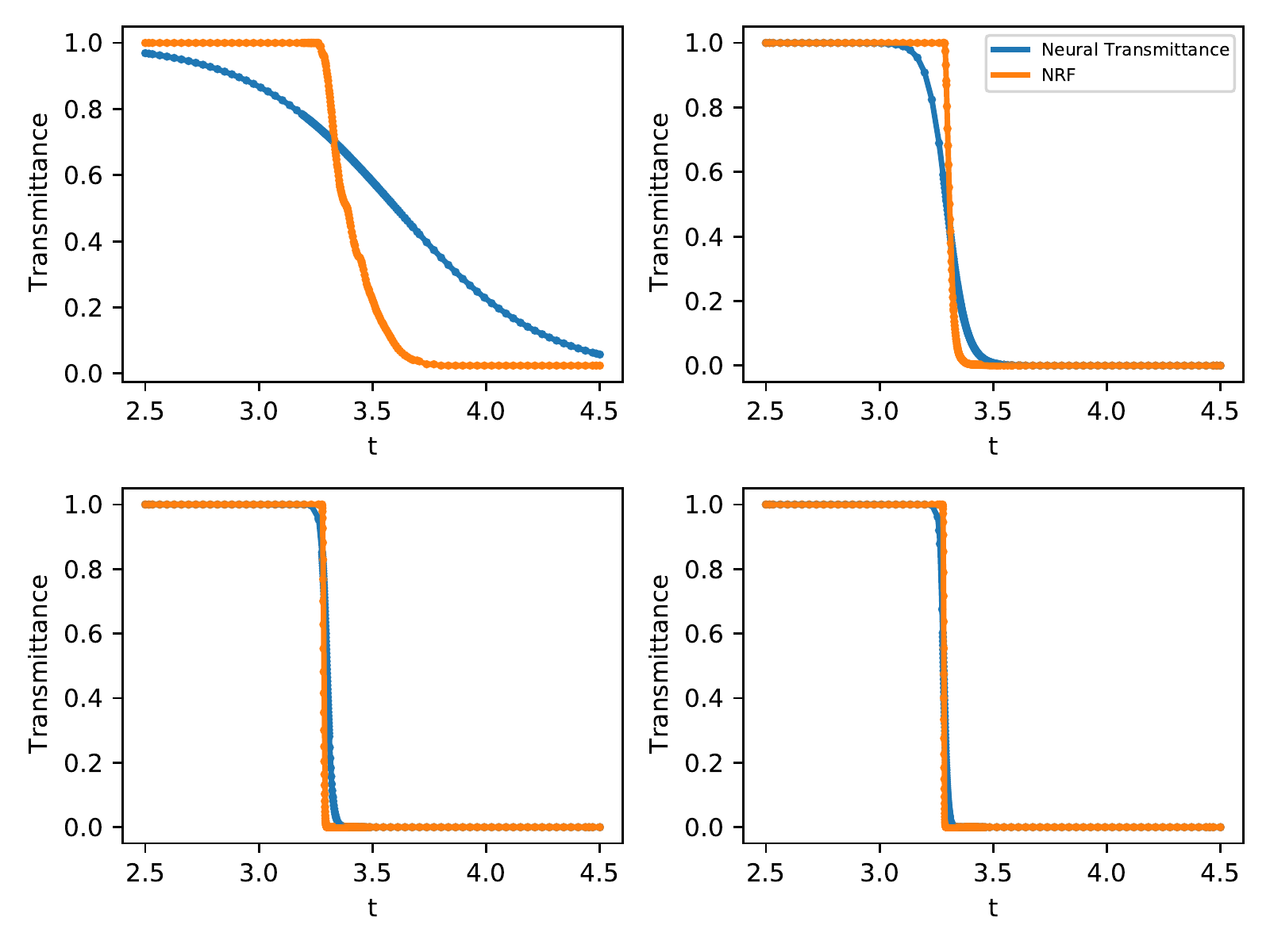}\\
  (d)
\end{minipage}

    \hfill    \caption{Numerically computed transmittance (red curve) is smooth
    in the early stages of training (a,b and c) and gradually converges to 
    nearly a step function (d). Neural transmittance function (blue curve) fits well to the numerically
    computed transmittance, while this function is simultaneously trained with those
    of NRF (training details explained in Section 4.1 in the main paper).
    (a),(b),(c) and (d) are respectively generated after 6000, 12000, 30000
    and 670000 iterations. }
\label{fig:transmittance}
\end{figure*}

\textbf{Neural Transmittance.}
Our method is based on the assumption that, objects in the scene are opaque.
More specifically, the transmittance function for an opaque scene can be well
modelled by a logistic function. Our proposed Neural Transmittance Function (Equation~7
in the main paper) estimates the transmittance along a ray by two parameters namely slope $a$ and center
$b$. We compare the neural transmittance to the transmittance
function computed by NRF~\cite{bi2020neural} in Figure~\ref{fig:transmittance}.
NRF~\cite{bi2020neural} computes the transmittance by the volume density values
queried from a neural network. These queries are made for the sampled points along a ray
in the ray marching step. Then, these values are used for numeric integration.
While in the Neural Transmittance Function, we simply estimate the transmittance
values for the same points by the prediction of function parameters $a$ and $b$.
These parameters allow us to compute the transmittance without ray marching.
In Figure~\ref{fig:transmittance} we compare our Neural Transmittance Function
to the transmittance computed by NRF~\cite{bi2020neural} in different stages
of training. In the early stages and before convergence the both functions are smooth.
In the late stages, the both functions look like a logistic function with large
slope. This figure is a plot of the transmittance values of points along a ray
in different stages of optimization. The ray is chosen from a training image of the
Sitting Buddha scene which passes from the geometry.

\begin{figure*}
  \centering
  \begin{minipage}[b]{0.50\textwidth}
    \centering
    \includegraphics[width=\textwidth]{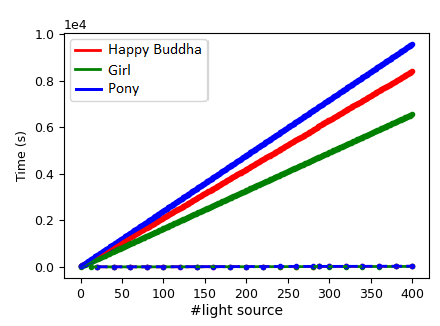}
  \end{minipage}
  
  \hfill    \caption{Runtime of our method (broken lines) compared to NRF (solid lines)
  for evaluating transmittance map and transmittance volume \cite{bi2020neural}.
  Transmittance map is faster to compute compared to transmittance volume for 3 different
  scenes each with a unique image resolution. The vertical axis is runtime in seconds and
  the horizontal axis is the resolution of environment maps in pixels. The lines corresponding
  to our method almost overlay each other on this plot.}
\label{fig:runtime_components}
\end{figure*}

\textbf{Runtime and complexity.}
Relighting a scene with NRF is a time consuming process especially for the case of
environment map rendering. We compare the time-complexity of our precomputation
method to that of Bi et al.~\cite{bi2020neural} in Table~\ref{tbl:complexity}.
We also compare the runtime of the precomputation step for our method
compared to NRF in Figure~\ref{fig:runtime_components}. Our method is significantly
faster for various scenes.

\begin{figure*}
  \centering
  \begin{minipage}[b]{0.3\textwidth}
    \centering
    \includegraphics[width=\textwidth]{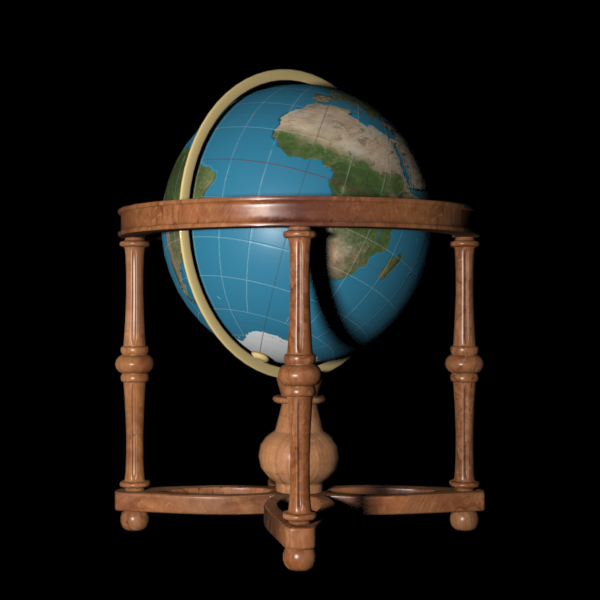} \\
        (a)
  \end{minipage}
  \begin{minipage}[b]{0.300\textwidth}
    \centering
      \includegraphics[width=\textwidth]{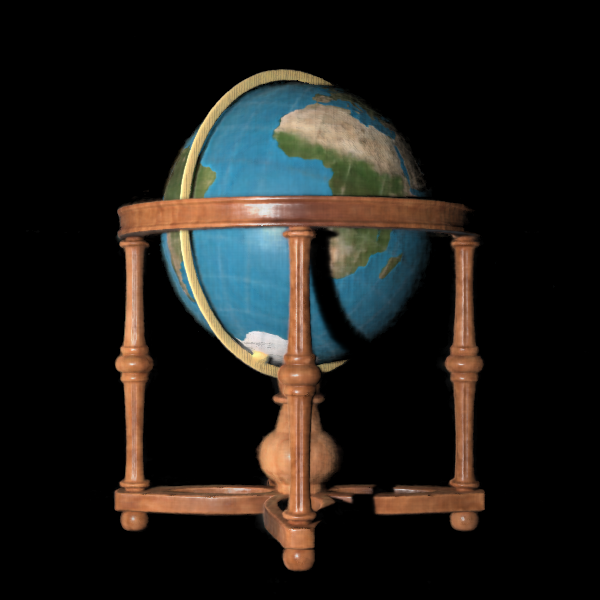} \\
      (b)
    \end{minipage}
    \begin{minipage}[b]{0.300\textwidth}
      \centering
          \includegraphics[width=\textwidth]{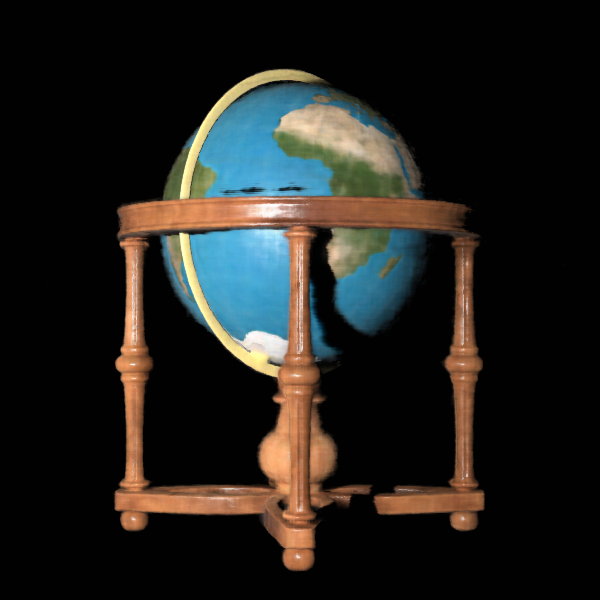} \\
          (c)
      \end{minipage}

      \begin{minipage}[b]{0.300\textwidth}
        \centering
          \includegraphics[width=\textwidth]{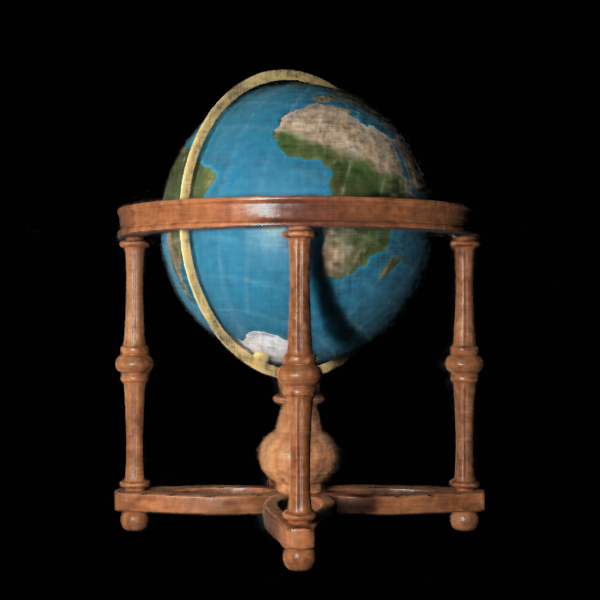} \\
          (d)
      \end{minipage}
      \begin{minipage}[b]{0.300\textwidth}
        \centering
          \includegraphics[width=\textwidth]{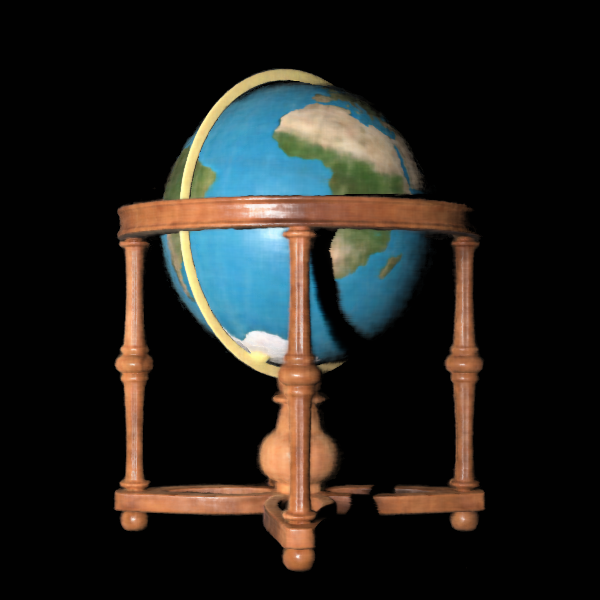} \\
          (e)
      \end{minipage}

  \hfill    \caption{We can relight the scene with our method (e)
  faster than NRF (b) and with less artifacts compared to NeRV
  visibility function (d). Our augmentation approach removes the artifacts
  caused by overfitting of neural transmittance on input images (c).}
\label{fig:ablation}
\end{figure*}

\textbf{Ablation.}
We show the effect of our augmentation method in Figure~\ref{fig:ablation}.
Relighting the globe scene without augmentation leads to clear artifacts.

\textbf{Visual comparison.} We visually compare our method with Bi et al.~\cite{bi2020neural}
in Figure~\ref{fig:ablation}. This figure includes relighting real and synthetic
scenes under point light and an environment map.

\begin{table}
    \small
    \centering
    \begin{tabular}{|c|c|c|c|} \hline
      \text{Naive}     & \text{NRF}      & \text{Ours AG}& \text{Ours TM} \\ \hline 
      $l \cdot n^2 \cdot c$        & $l \cdot m \cdot n$       & $l \cdot m \cdot n$  & $l \cdot m$          \\ \hline 
    \end{tabular}
    \caption{Time-complexity of relighting a scene with an environment map.
    $l, m, n$ and $c$ are respectively the number of pixels on an environment map,
    pixels on a virtual camera on each light source, samples on a ray and pixels on the camera.}
    \label{tbl:complexity}
\end{table}

\bibliography{references}